\documentclass[journal]{IEEEtran}
\usepackage{amsmath,amsfonts}
\usepackage{algorithmic}
\usepackage{algorithm}
\usepackage{array}
\usepackage[caption=false,font=normalsize,labelfont=sf,textfont=sf]{subfig}
\usepackage{textcomp}
\usepackage{stfloats}
\usepackage{url}
\usepackage{verbatim}
\usepackage{graphicx}
\usepackage{cite}

\usepackage{booktabs}                                   % for nice tables
\usepackage{multirow}                                   % for table cell taking multiple rows
\usepackage{tablefootnote}                              % for footnote in table
\usepackage[symbol]{footmisc}                           % for footnote symbol
\usepackage{amsmath,amssymb}                            % for \checkmark, etc.
\usepackage{xcolor}                                     % for colors
\usepackage{enumitem}                                   % for spaces in \itemize
\usepackage{marvosym}                                   % for \Letter used in author list
\usepackage{ifsym}                                      % for \Letter used in author list
\usepackage{stfloats}
\usepackage{makecell}                                   % for table cell with texts in multiple rows

\usepackage{gensymb}

\newcommand{\field}[1]{\mathbb{#1}}
\newcommand{\R}{\field{R}}                              % real domain
\newcommand{\Loss}{\mathcal{L}}                    % loss
\newcommand{\vn}{\mathbf{n}}
\newcommand{\vz}{\mathbf{v}_{\text{z}}}
\newcommand{\zz}{\mathbf{z}}

\newcommand{\dxx}[1]{\mathbf{d}_{\text{#1}}}

\newcommand{\dr}{\dxx{raw}}
\newcommand{\dpred}{\dxx{pred}}
\newcommand{\dgt}{\dxx{gt}}
\newcommand{\dl}{\dxx{l}}
\newcommand{\df}{\dxx{f}}
\newcommand{\cl}{\mathbf{c}_{\text{l}}}
\newcommand{\cf}{\mathbf{c}_{\text{f}}}
\newcommand{\rr}{\mathbf{r}}
\newcommand{\fr}{\mathbf{f}_\rr}

\newcommand{\pp}{\mathbf{p}}

\newcommand{\PPRED}{\mathcal{P}_\text{pred}}
\newcommand{\PGT}{\mathcal{P}_\text{gt}}

\newcommand{\RR}{\mathcal{R}}
\newcommand{\TT}{\mathcal{T}}
\newcommand{\RS}{\mathcal{R^*}}
\newcommand{\TS}{\mathcal{T^*}}

\newcommand{\T}{^{\top}}                                % transpose

\newcommand{\para}[1]{\subsubsection{#1}}
\def\wada{\mbox{W-AdaIN}}
\def\mcn{\mbox{MCN}}
\def\rdfg{\mbox{RDF-GAN}}
\def\rdfcg{\mbox{RDFC-GAN}}

\newcommand{\etal}{\textit{et al}.}

\begin{document}
\title{{\rdfcg}: RGB-Depth Fusion CycleGAN\\for Indoor Depth Completion}

\author{
  Haowen~Wang$^*$,~\IEEEmembership{Student Member,~IEEE},
  Zhengping~Che$^*$,~\IEEEmembership{Member,~IEEE},
  Yufan~Yang,
  Mingyuan~Wang,
  \\
  
  Zhiyuan~Xu,~\IEEEmembership{Member,~IEEE},
  Xiuquan~Qiao\textsuperscript{\textrm{\Letter}},
  Mengshi~Qi,
  Feifei~Feng,
  and Jian~Tang\textsuperscript{\textrm{\Letter}},~\IEEEmembership{Fellow,~IEEE}% <-this % stops an unwanted space
\thanks{
  Haowen Wang, Yufan Yang, Mingyuan Wang, and Xiuquan Qiao are with
  State Key Laboratory of Networking and Switching Technology, Beijing University of Posts and Telecommunications, China.
  E-mail:  \{hw.wang, yufanyang, wmingyuan, qiaoxq\}@bupt.edu.cn
}% <-this % stops an unwanted space
\thanks{
  Zhengping Che, Zhiyuan Xu, Feifei Feng, and Jian Tang are with Midea Group, China.
  E-mail:  chezhengping@gmail.com and \{chezp, xuzy70, feifei.feng, tangjian22\}@midea.com
}% <-this % stops an unwanted space
\thanks{
  Mengshi Qi is with School of Computer Science, Beijing University of Posts and Telecommunications, China.
  E-mail:  qms@bupt.edu.cn
}% <-this % stops an unwanted space
\thanks{
  This work was done during Haowen Wang's internship at Midea Group.
  A preliminary version~\cite{wang2022rgb} of this paper was presented at
  the 35th IEEE/CVF Conference on Computer Vision and Pattern Recognition (CVPR), 2022.
}% <-this % stops an unwanted space
\thanks{
  $^*$First Authors: Haowen Wang and Zhengping Che contributed equally.
}% <-this % stops an unwanted space
\thanks{
  \textsuperscript{\textrm{\Letter}}Corresponding Authors: Xiuquan Qiao and Jian Tang.
}% <-this % stops an unwanted space
% \thanks{Manuscript received April 19, 2005; revised August 26, 2015.}
}

\markboth{{\rdfcg}: RGB-Depth Fusion CycleGAN for Indoor Depth Completion}%
{{\rdfcg}: RGB-Depth Fusion CycleGAN for Indoor Depth Completion}

\maketitle

\begin{abstract}
	Raw depth images captured in indoor scenarios frequently exhibit extensive missing values due to the inherent limitations of the sensors and environments.
For example, transparent materials frequently elude detection by depth sensors; surfaces may introduce measurement inaccuracies due to their polished textures, extended distances, and oblique incidence angles from the sensor.
The presence of incomplete depth maps imposes significant challenges for subsequent vision applications, prompting the development of numerous depth completion techniques to mitigate this problem.
Numerous methods excel at reconstructing dense depth maps from sparse samples, but they often falter when faced with extensive contiguous regions of missing depth values, a prevalent and critical challenge in indoor environments.
To overcome these challenges, we design a novel two-branch \mbox{end-to-end} fusion network named {\rdfcg}, which takes a pair of RGB and incomplete depth images as input to predict a dense and completed depth map.
The first branch employs an encoder-decoder structure,
by adhering to the Manhattan world assumption and utilizing normal maps from RGB-D information as guidance,
to regress the local dense depth values from the raw depth map.
The other branch applies an RGB-depth fusion CycleGAN, adept at translating RGB imagery into detailed, textured depth maps while ensuring high fidelity through cycle consistency.
We fuse the two branches via adaptive fusion modules named {\wada} and train the model with the help of pseudo depth maps.
Comprehensive evaluations on NYU-Depth V2 and SUN RGB-D datasets show that our method significantly enhances depth completion performance particularly in realistic indoor settings.

\end{abstract}


\begin{figure}[t]
  \centering
  \includegraphics[width=0.99\linewidth]{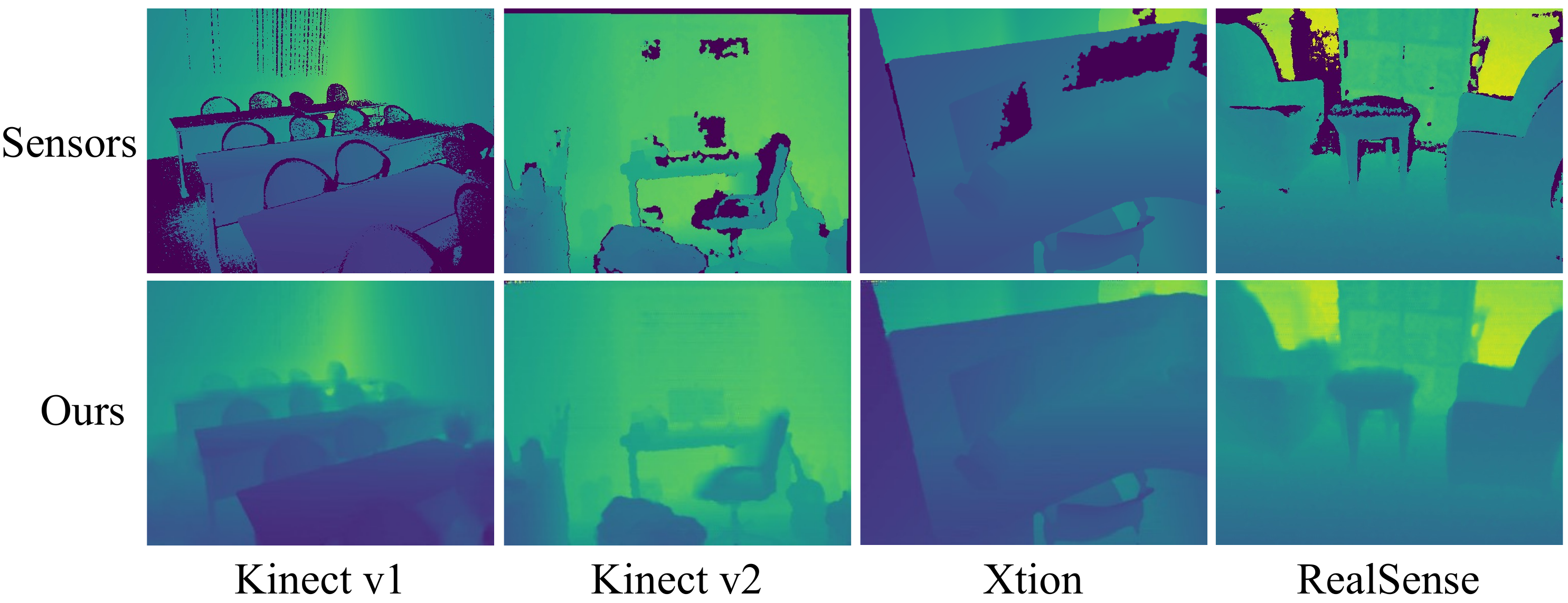}
  \caption{
    Showcases of the raw depth maps (top) in indoor scenarios collected by different sensors from the {SUN RGB-D} dataset~\cite{song2015sun} and the corresponding depth completion results (bottom) of our method.
  }
  \label{fig:sunrgb}
\end{figure}

\section{Introduction}\label{sec:introduction}

\IEEEPARstart{D}EPTH map, also known as depth image, as a reliable representation of 3D spatial information, has been widely used in many vision applications including augmented reality, indoor navigation, and 3D reconstruction tasks~\cite{Fu_2018_CVPR,li2019vision,zhao2020pointar}.
However, most existing commercial depth sensors (e.g., Kinect~\cite{kinect}, RealSense~\cite{Keselman_2017_CVPR_Workshops}, and Xtion~\cite{xtion}) for indoor spatial perception are not powerful enough to generate a precise and lossless depth map, as shown in the top row of Fig.~\ref{fig:sunrgb}.
The prevalence of incomplete depth maps in indoor settings largely stems from inherent sensor limitations and the intrinsic properties of the scene, and these holes significantly affect the performance of downstream tasks on the depth maps.
For instance, laser scanners and structured-light sensors frequently fail to detect surfaces like windows and glass, since the light passes straight through these transparent materials rather than reflecting back.
Likewise, smooth surfaces such as ceilings and walls can reflect or absorb light, leading to gaps in the depth data.
Distance extremes and acute angles of incidence relative to the sensor's orientation further contribute to these incomplete measurements, underscoring the need for sophisticated depth completion techniques to address these deficiencies.

To mitigate the challenges presented by imperfect depth maps, a multitude of approaches, collectively known as \textit{depth completion},
have been developed to reconstruct comprehensive depth maps from their incomplete counterparts.
Depth completion often involves utilizing a concurrent pair of raw depth and RGB images, obtained from a single depth-sensing device, to fill in the missing depth information and refine the depth map's accuracy.
Recent studies have produced significant progress in depth completion tasks with convolutional neural networks~(CNNs)~\cite{ma2018sparse,cheng2018depth,qiu2019deeplidar,huang2019indoor,lee2019depth,park2020non}.
Ma and Karaman~\cite{ma2018sparse} introduced an encoder-decoder network to directly regress the dense depth map from a sparse depth map and an RGB image.
The method has shown great progress compared to conventional algorithms~\cite{saxena2005learning,liu2014discrete,yang2014stereo},
but its outputs are often too blurry because of the lack of captured local information.

To generate a more refined completed depth map, lots of works have recently arisen, which can be divided into two groups with different optimization methods.
The first group of works~\cite{liu2017learning,cheng2018depth,park2020non} learn affinities for relative pixels and iteratively refine depth predictions, which highly rely on the accuracy of the raw global depth map and suffer the inference inefficiency.
Other works~\cite{huang2019indoor,lee2019depth,qiu2019deeplidar,lee2021depth} analyze the geometric characteristic and adjust the feature network structure accordingly, for instance, by estimating the surface normal or projecting depth into discrete planes.
Meanwhile, existing methods use the RGB image as guidance or auxiliary information.
For example, based on statistics extracted from image-depth pairs, a common prior that depth discontinuities are largely aligned with the edges in the image has been widely adopted~\cite{zhong2019deep,ding2023rethinking}.
However, methods that adequately investigate deeper correlation between RGB semantic features and depth maps are still in great demand.
Also, the model parameters may not be efficiently generalized to different scenes, as few methods deeply consider the textural and contextual information, and the model parameters may not be efficiently generalized to different scenes.

It is also worth noting that depth completion in indoor environments, due to its special properties, has not been well-addressed by existing depth completion methods.
Prevailing depth completion approaches~\cite{huang2019indoor,lee2019depth,qiu2019deeplidar,lee2021depth,lin2022dynamic} emphasize intricate adaptive propagation structures for local pixels, which may fail in dealing with large invalid depth maps that are prevalent in indoor scenes.
Furthermore, it is common that man-made houses follow regular geometric structures, such as mutual perpendicular-oriented walls, floors, and ceilings.
This domain knowledge, usually referred to as Manhattan world assumption~\cite{coughlan2000manhattan}, can help people easily tell invalid and unreasonable depth estimation results and has been properly used in SLAM~\cite{yunus2021manhattanslam}, monocular depth estimation~\cite{Li_2021_ICCV}, and 3D reconstruction~\cite{guo2022neural}.
However, to effectively incorporate this structural regularity in depth completion methods, especially with the fusion of RGB and depth images, is unexplored.

\begin{figure}[t]
  \centering
  \includegraphics[width=.95\linewidth]{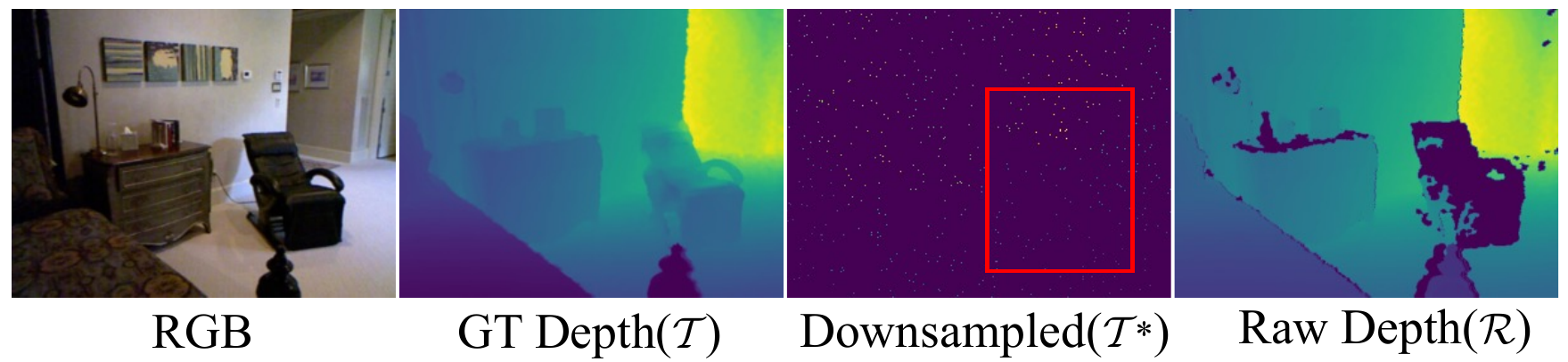}\\
  \includegraphics[width=.95\linewidth]{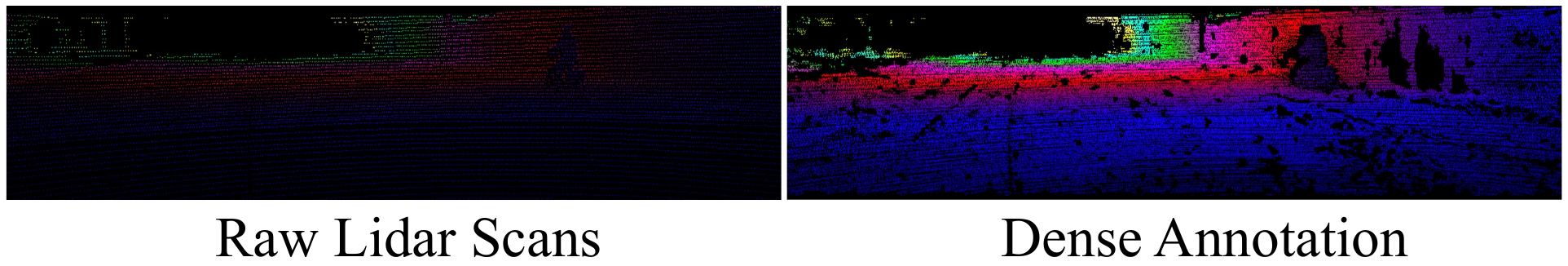}
  \caption{
    Depth data visualizations of indoor RGB-Depth sensor data (top, NYU-Depth~V2) and outdoor Lidar scan data (bottom, KITTI).
    The downsampled data ($\mathcal{T^*}$) is 500 pixels randomly and uniformly sampled from the ground-truth (GT) depth data ($\mathcal{T}$),
    which contains ground truth depth values (e.g., in the red box) that do not exist in the raw depth data ($\mathcal{R}$).
  }
  \label{fig:setting}
\end{figure}

More remarkably, most existing methods~\cite{ma2018sparse,cheng2018depth,lee2021depth} only consider completing sparse depth images and uniformly randomly sample a certain number of valid pixels from the \text{raw or complete} dense depth image as the input for training and evaluation.
While such downsampling setting mimics well the task of outdoor depth completion from \textit{raw Lidar scans} to dense annotations (as shown in the bottom row of Fig.~\ref{fig:setting}), it is improper for indoor RGB-depth sensor data,
since the sampled patterns are quite different from the real missing patterns in indoor scenes, such as large missing regions and semantical missing patterns.
Specifically, as shown in the top row of Fig.~\ref{fig:setting}, the raw depth map captured by indoor depth sensors is dense and continuous, which is quite different from the sparse pattern of downsampled input.
Meanwhile, the downsampled input leaks the ground truth depth values in the mimicked missing regions to the completion models, leading to flawed evaluations.
Thus, it is unclear whether the successful methods in uniformly sparse depth map settings still win in indoor depth completion tasks.
This should be addressed by reasonable training strategies and comprehensive evaluation settings specifically designed for indoor scenarios.

To solve these problems in indoor depth completion, we propose a novel \mbox{two-branch} end-to-end network to generate a completed dense depth map for indoor environments.
On the one hand,
inspired by a series of generative adversarial networks~(GANs)~\cite{mirza2014conditional,kim2017learning,karras2019style,ma2019fusiongan} including \mbox{CycleGANs}~\cite{isola2017image,CycleGAN2017} that can effectively capture and exploit texture style information,
we propose an \textit{RGB-depth Fusion CycleGAN}~(\rdfcg) branch for fusing an RGB image and a depth map.
The cycle consistency loss of CycleGAN is crucial for preserving essential features and textures, securing detailed and authentic depth maps that faithfully reflect the original scene's structure.
On the other hand, we design a \textit{Manhattan-Constraint Network}~(\mcn) branch that leverages the Manhattan world assumption in a generated normal map to guide the depth completion in indoor scenes.
In order to connect these two branches and refine the estimated depth, we introduce \textit{weighted adaptive instance normalization}~(\wada) modules and uses a \textit{confidence fusion head} to conclude the final results.
In addition, we produce pseudo depth maps for training by sampling raw depth images in accordance with the indoor depth missing characteristics.

Our main contributions are summarized as the following:
\begin{itemize}[itemsep=1pt,topsep=1pt,parsep=1pt,leftmargin=10pt]
    \item We propose a novel end-to-end network named {\rdfcg} that effectively fuses a raw depth map and an RGB image to produce a complete dense depth map in indoor scenarios.
    \item We design the Manhattan-constraint network utilizing the geometry properties of indoor scenes, which effectively introduce smoother depth value constraints and further boosts the performance of {\rdfcg}.
    \item We elaborate the definition and training usage of pseudo depth maps that mimic indoor raw depth missing patterns and can improve depth completion model performance.
    \item We show that our proposed method achieves state-of-the-art performance on {NYU-Depth V2} and {SUN RGB-D} for depth completion with comprehensive evaluation metrics and prove its effectiveness in improving downstream task performance such as object detection.
\end{itemize}

\begin{figure*}[b!]
  \centering
  \includegraphics[width=.95\linewidth]{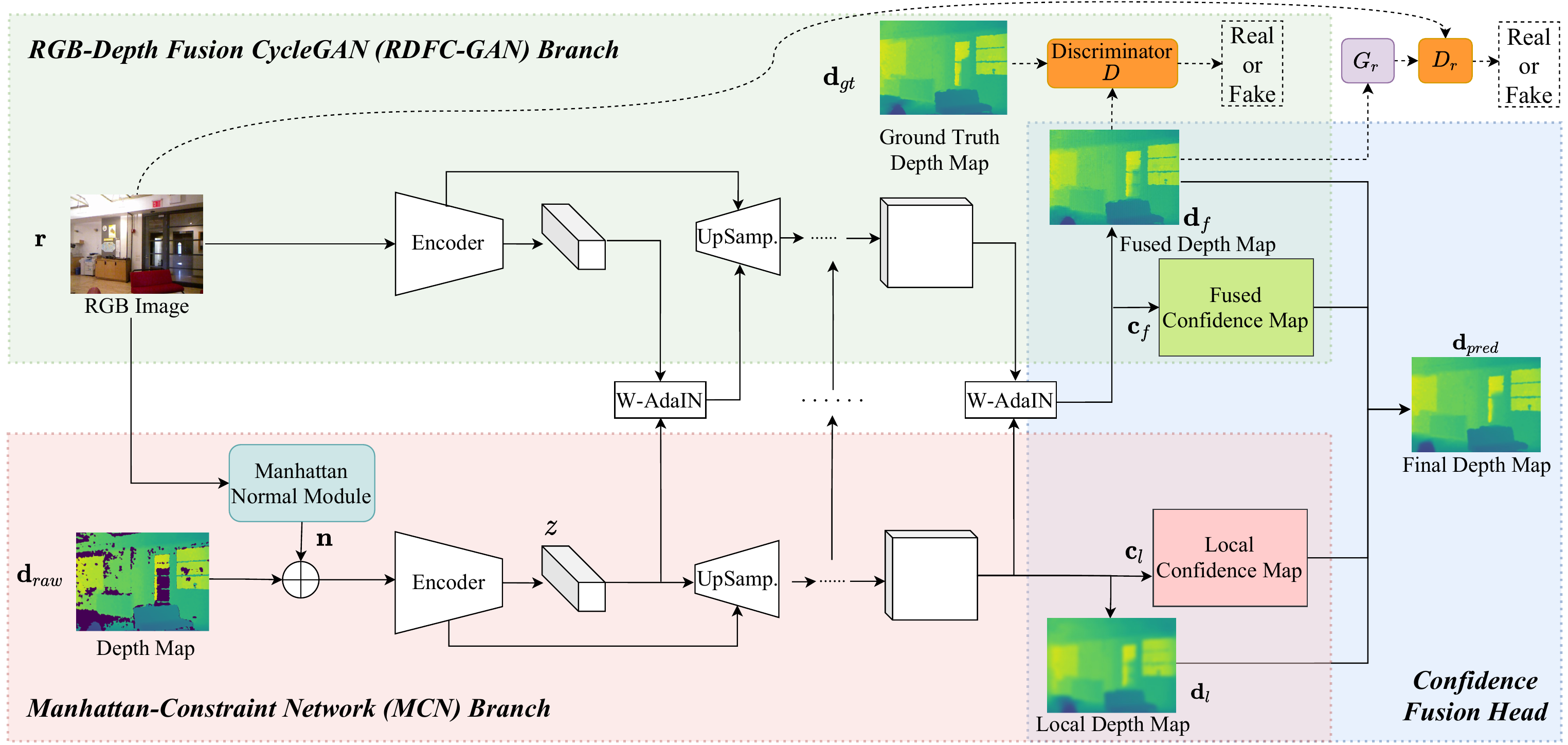}
  \caption{
    The overview of the proposed end-to-end depth completion method (\rdfcg).
    Compared to the preliminary model \rdfg~\cite{wang2022rgb}, the Manhattan normal module and the CycleGAN are the main structural improvements in {\rdfcg}.
  }
  \label{fig:architect}
\end{figure*} 
\section{Related Work}
\para{Depth Completion with Deep Learning}
Recent works have extensively applied deep neural networks for depth completion tasks with remarkable improvements.
Ma and Karaman~\cite{ma2018sparse} used a CNN encoder-decoder to predict the full-resolution depth image from a set of depth samples and RGB images.
On this basis, several methods~\cite{qiu2019deeplidar,huang2019indoor,depthcompletion_gan,lee2021depth,yan2022rignet} incorporating additional representations or auxiliary outputs have been proposed.
Qiu~\etal~\cite{qiu2019deeplidar} produced dense depth using the surface normal as the intermediate representation.
Imran~\etal~\cite{imran2019depth} introduced the depth coefficients to address the challenge of depth smearing between objects.
Lee~\etal~\cite{lee2021depth} factorized the depth regression problem into a combination of discrete depth plane classification and plane-by-plane residual regression.
Chen~\etal~\cite{chen2022depth} converted the depth map to the point clouds and used geometry-aware embedding to fill in missing depth information.
Another series of methods~\cite{cheng2018depth,van2019sparse,park2020non,lin2022dynamic} have introduced new network structures to depth completion tasks.
Cheng~\etal~\cite{cheng2018depth} proposed the convolutional spatial propagation network~(CSPN) and generated the long-range context through a recurrent operation. 
Li~\etal~\cite{li2020multi} introduced a multi-scale guided cascade hourglass network to capture structures at different levels.
Senushkin~\etal~\cite{senushkin2020decoder} controlled the depth decoding for different regions via spatially-adaptive denormalization blocks.
NLSPN~\cite{park2020non} improved CSPN by non-local spatial and global propagations.
DySPN~\cite{lin2022dynamic} and GraphCSPN~\cite{liu2022graphcspn} further enhance the performance of sparse depth completion tasks.
In this work, to build our depth completion model, we both include new representations and extend the network structure.

\para{RGB-D Fusion}
The fusion of both RGB and depth data (a.k.a., the RGB-D fusion) is essential in many tasks such as semantic segmentation\cite{chen2020bi,cao2021shapeconv,zhou2022canet}, scene reconstruction~\cite{bozic2020deepdeform,wu2021scenegraphfusion,azinovic2022neural}, and navigation~\cite{karkus2021differentiable,teed2021droid,FANG2021107822}.
While early works~\cite{maire2016affinity,ma2018sparse} only concatenate aligned pixels from RGB and depth features, more effective RGB-D fusion methods have been proposed recently.
Cheng~\etal~\cite{cheng2017locality} designed a gated fusion layer to learn different weights of each modality in different scenes.
Park~\etal~\cite{park2017rdfnet} fused multi-level RGB-D features in a very deep network through residual learning.
Du~\etal~\cite{du2019translate} proposed a cross-modal translate network to represent the complementary information and enhance the discrimination of extracted features.
Our {\rdfcg} uses a two-branch structure and progressively deploys the {\wada} modules to better capture and fuse RGB and depth features.

\para{Generative Adversarial Networks}
Generative adversarial networks~(GANs)~\cite{mirza2014conditional} have achieved great success in a variety of image generation tasks such as style transfer~\cite{lin2020gan,emami2020spa,li2021image}, realistic image generation~\cite{liu2021pd,afifi2021histogan}, and image synthesis~\cite{zhan2019spatial,chan2021pi}.
CycleGAN~\cite{zhu2017unpaired} maintains the inherent features of the source domain while translating to the target domain through the cycle consistency.
Karras~\etal~\cite{karras2019style} introduced a style-based GAN to embed the latent code into a latent space to affect the variations of generated images.
This work uses a CycleGAN-based structure, extending the preliminary \mbox{GAN-based} one~\cite{wang2022rgb}, to generate completed depth maps.

\para{Indoor Structural Regularities}
The Manhattan world assumption~\cite{coughlan2000manhattan} takes advantage of the prevalent orthogonal directions in human-made environments, thereby streamlining tasks by diminishing the intricacy of the associated structures.
An increasing number of methods~\cite{zou2018layoutnet,lin2019floorplan,guo2022neural} leverage it for indoor vision tasks.
For example, in 3D room layout estimation tasks, some works~\cite{zou2018layoutnet,lin2019floorplan,yang2019dula,pintore2020atlantanet,zou2021manhattan} simplify task complexity by transforming corner points or junctions into intersecting vertical planes.
Li~\etal~\cite{Li_2021_ICCV} exploited the inherent structural regularities to improve monocular depth estimation.
In 3D scene understanding and reconstruction, Manhattan world assumption has been integrated as connections between 3D scenes and 2D images~\cite{zhang2014panocontext} or as additional constraints~\cite{furukawa2009manhattan,li2016manhattan,guo2022neural}.
In this work, we introduce the Manhattan world assumption into depth completion tasks for the first time. 

\section{Method}
In this section, we describe the proposed depth completion method, as shown in Fig.~\ref{fig:architect}.
The model takes a raw (noisy and possibly incomplete) depth map $\dr \in \R^{H \times W \times 1}$ and its corresponding RGB image $\rr \in \R^{H \times W \times 3}$ as the input,
and outputs the completed and refined dense depth map estimation (a.k.a, final depth map) $\dpred \in \R^{H \times W \times 1}$ to be close to the ground truth depth map $\dgt \in \R^{H \times W \times 1}$,
where $H$ and $W$ are the height and the width of the depth map, respectively.

The model mainly consists of two branches:
a Manhattan-Constraint Network (\mcn) branch (Section~{\ref{sec:local_dc}}) and an RGB-depth Fusion CycleGAN (\rdfcg) branch (Section~{\ref{sec:gan_dc}}).
{\mcn} and {\rdfcg} take the depth map and the RGB image as the input, respectively, and produce their individual depth completion results.
To fuse the representations between the two branches,
a series of intermediate fusion modules called {\wada} (Section~{\ref{sec:connect_module}}) are deployed at different stages of the model.
Finally, a confidence fusion head (Section~{\ref{sec:confidence_fusion}}) combines the outputs of the two channels and provides more reliable and robust depth completion results.
Moreover, we introduce the training strategy with pseudo depth maps (Section~\ref{sec:pseudo depth map}) and describe the overall loss function for training (Section~\ref{sec:loss_function}).

\begin{figure}[b]
\centering
\includegraphics[width=0.95\linewidth]{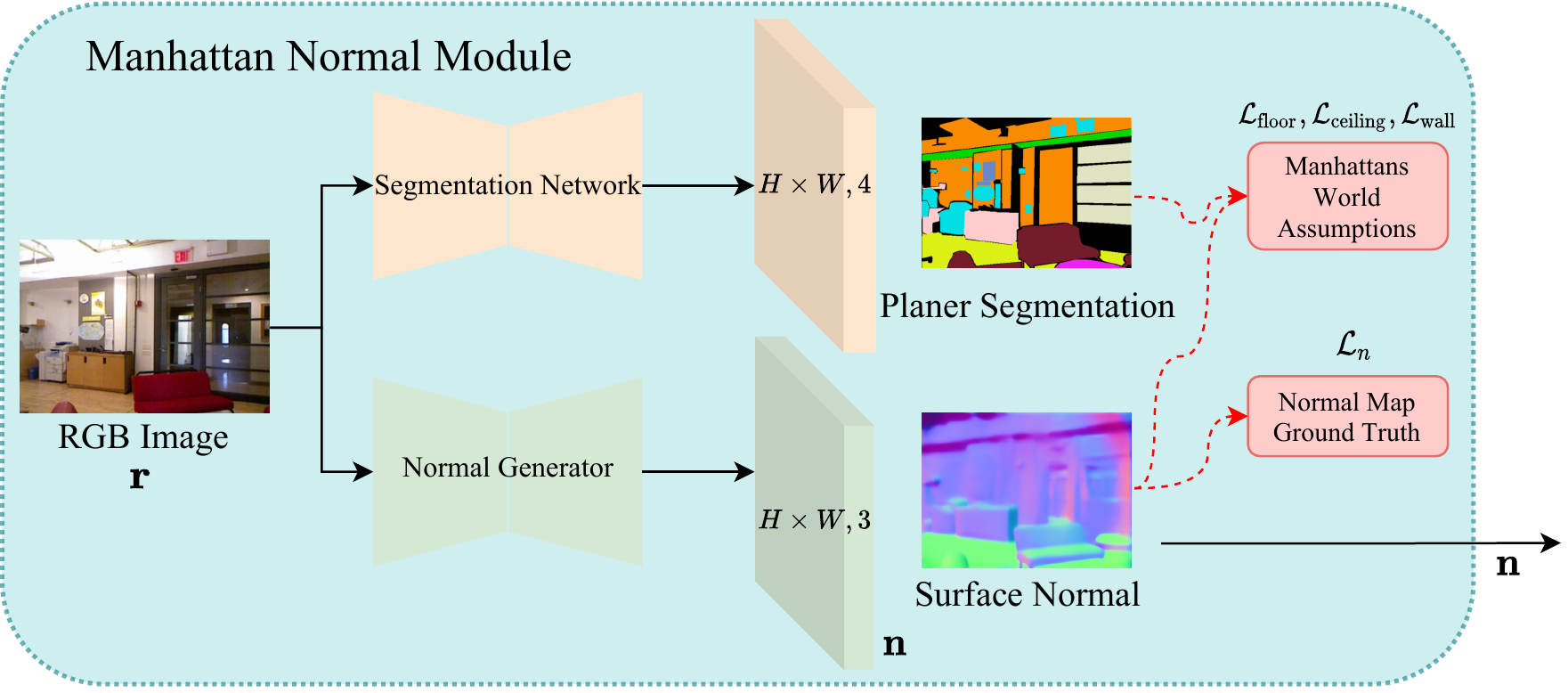}
\caption{An Illustration of the Manhattan normal module in the Manhattan-Constraint network (\mcn).}
\label{fig:local_dc}
\end{figure} 
\subsection{The Manhattan-Constraint Network (\mcn) Branch}
\label{sec:local_dc}
The first branch, Manhattan-Constraint Network (\mcn) branch, is composed of a Manhattan normal module and a convolutional encoder-decoder structure. 
As illustrated in the bottom-left part of Fig.~\ref{fig:architect},
this branch mainly relies on the raw depth map, as well as auxiliary from the RGB image, and outputs a dense local depth map $\dl \in \R^{H \times W \times 1}$ and a local confidence map $\cl \in \R^{H \times W \times 1}$.

\para{Manhattan Normal Module}
Depth prediction in coplanar regions can benefit from known surface normals~\cite{patil2022p3depth,xu2022multi}.
However, estimating surface normals in indoor scenes is challenging due to pervasive large untextured planes with consistent luminosity in rooms.
To address this, we design a Manhattan normal module to leverage the Manhattan World assumption~\cite{coughlan2000manhattan} that most surfaces in indoor scenes are usually orthogonal and aligned with three dominant directions, which is shown in Fig.~\ref{fig:local_dc}.
On one hand, we employ a pre-trained segmentation network~\cite{zhao2017pyramid} to identify floor, ceiling, and wall regions in the RGB scene.
Also, we use a U-Net~\cite{bae2021estimating} as a normal generator to generate a normal map that can both approximate the ground truth and follow the Manhattan assumption.

Specifically, for all predicted normal vectors $\vn_p \in \R^3$ where $p$ refers to any pixel,
we optimize the cosine similarity loss $\Loss_{\vn}$ between the predicted normal vectors and the ground-truth normal map by
\begin{equation}
  \Loss_{\vn} = -\frac{1}{H W} \sum_{p} \frac{\vn_p \cdot \vn^*_p}{\|\vn_p\| \cdot \|\vn^*_p\|},
\end{equation}
where $\vn^*$ is the ground truth.
For planar regions, we incorporate the information from segmentation results (i.e., whether each pixel $p$ belongs to the floor, ceiling, wall, or none)
and ensure the normals to be consistent with plane physical orientations.
For example, we enforce all floor points to be upward perpendicular oriented by
\begin{align}
  \Loss_\text{floor} & = - \frac{1}{ \sum_p \mathcal{I}(p \in \textrm{floor})}
  \sum_p \frac{\vn_p \cdot \vz }{\|\vn_p\|}\mathcal{I}(p \in \textrm{floor}),
\end{align}
where $\vz = (0, 0, 1)$ is the upward perpendicular unit normal vector, and $\mathcal{I}(\cdot)$ is the indicator function.
Similarly, the ceiling points and wall points are constrained to point downward and horizontally, respectively, and we have
\begin{align}
  \Loss_\text{ceiling} & = \frac{1}{ \sum_p \mathcal{I}(p \in \textrm{ceiling})}
  \sum_p \frac{\vn_p \cdot \vz }{\|\vn_p\|}\mathcal{I}(p \in \textrm{ceiling}), \\
  \Loss_\text{wall} & = \frac{1}{ \sum_p \mathcal{I}(p \in \textrm{wall})}
  \sum_p \frac{| \vn_p \cdot \vz |}{\|\vn_p\|}\mathcal{I}(p \in \textrm{wall}).
\end{align}

In summary, the loss for the Manhattan normal module is
\begin{equation}
  \Loss_\text{MNM} =  \Loss_{\vn} +  \Loss_\text{floor} +  \Loss_\text{ceiling} + \Loss_\text{wall}.
\end{equation}

\begin{figure}[b]
\centering
\includegraphics[width=0.95\linewidth]{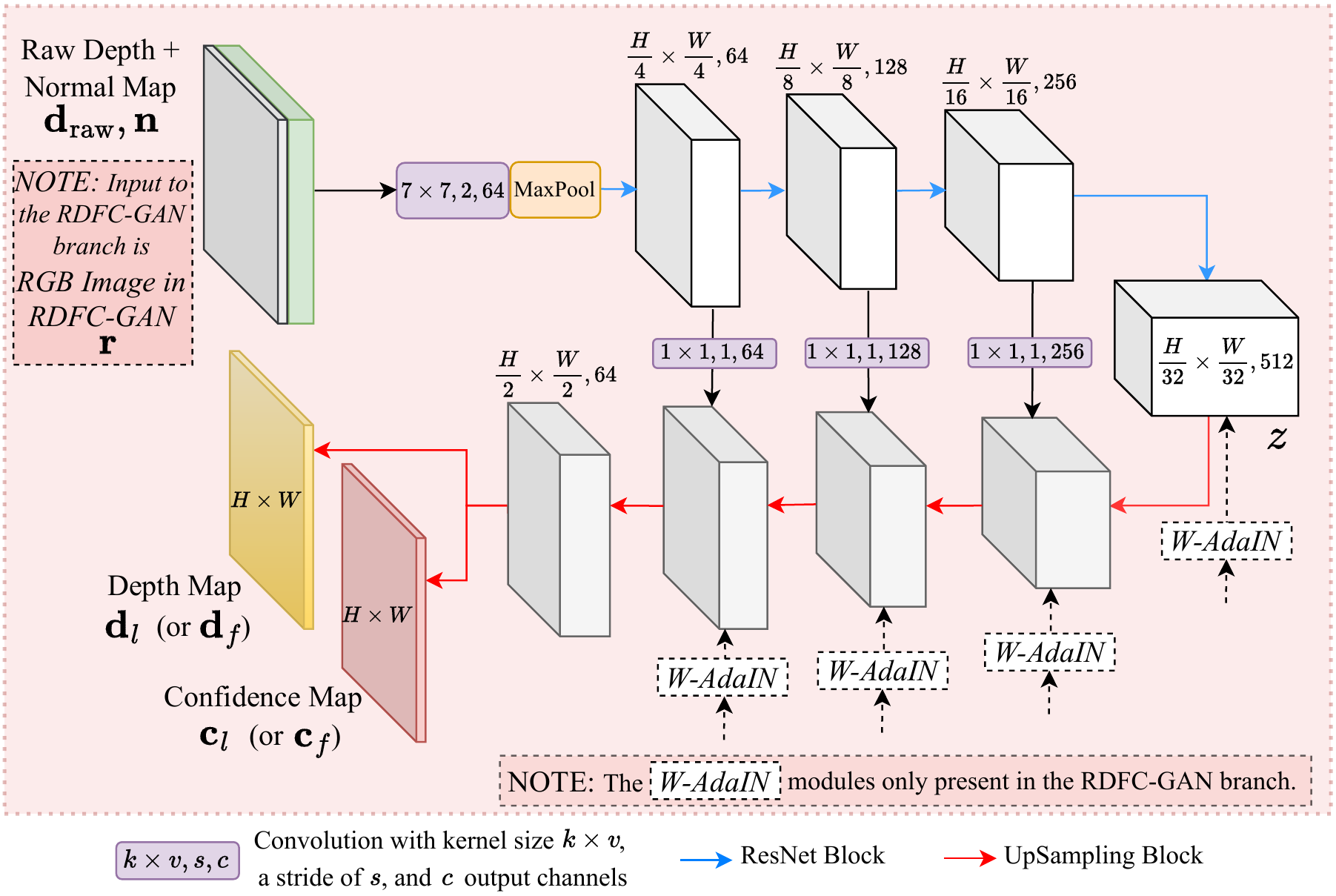}
\caption{An Illustration of the encoder-decoder structure in the two branches.}
\label{fig:encoder-decoder}
\end{figure}

\para{Encoder-Decoder Structure}
The output of the Manhattan normal module (i.e., a three-channel map $\vn \in \R^{H \times W \times 3}$) is concatenated with the one-channel raw depth image $\dr$ to form the input to an encoder-decoder.
The encoder-decoder of {\mcn}, as shown in Fig.~\ref{fig:encoder-decoder}, is based on ResNet-18~\cite{he2016deep} and pre-trained on the ImageNet dataset~\cite{deng2009imagenet}.
Given this input, the encoding stage downsamples the feature size by 32 times and expands the feature dimension to 512.
The encoder learns the mapping from the depth map space to the depth latent space and produces $\zz \in \R^{\frac{H}{32} \times \frac{W}{32} \times 512}$ as the fused depth feature information. 
The decoding stage applies a set of upsampling blocks to increase the feature resolution with the skip connection from the encoder.
The output of the decoder is a local depth map and its corresponding local confidence map, which is the final output of the {\mcn} branch.

The overall loss for the {\mcn} branch $\Loss_{\text{MCN}}$ also includes the $L_1$ loss on the local depth map, i.e.,
\begin{align}
\Loss_{\text{MCN}} = \Loss_{\text{MNM}} + \lambda_\text{l} {\| \dl - \dgt \|}_1,
\end{align}
where $\lambda_\text{l}$ is the weight hyperparameter for the $L_1$ loss.

\subsection{The RGB-Depth Fusion CycleGAN (\rdfcg) Branch}
\label{sec:gan_dc}
To generate the fine-grained textured and dense depth map, we propose the second branch in our model, which is a \mbox{GAN-based} structure for RGB and depth image fusion, as illustrated in the top-left part of Fig.~\ref{fig:architect}.
Different from most existing fusion methods that directly concatenate inputs from different domains, our fusion model, inspired by the conditional and style GANs~\cite{mirza2014conditional,karras2019style},
a) uses the depth latent vector mapping from the incomplete depth image as the input and the RGB image as the condition to generate a dense fused depth prediction $\df \in \R^{H \times W \times 1}$ and a fused confidence map $\cf \in \R^{H \times W \times 1}$,
and b) uses a discriminator to distinguish the ground truth depth images from generated ones.

The generator $G(\cdot)$ has a similar structure as the encoder-decoder of {\mcn} shown in Fig.~\ref{fig:encoder-decoder}, except for the RGB-only input and the fusion with {\wada}.
Given the corresponding RGB image $\rr$ as the condition, the generator $G(\cdot)$ with the depth latent vector $\zz$ generates a fused dense depth map $\df$ and a fused confidence map $\cf$ for the scene.
The latent vector $\mathbf{z}$ from {\mcn} propagates the depth information to the RGB image using the proposed {\wada} described later in Section~\ref{sec:connect_module}.
We distinguish the fused depth map $\df$ and the real depth image $\dgt$ by the discriminator ${D(\cdot)}$, whose structure is based on PatchGAN~\cite{isola2017image}.

Besides the main GAN structure, to enhance the effects of texture information in generating depth maps, 
we form a structure of CycleGAN~\cite{zhu2017unpaired} with an auxiliary pair of generator $G_{\rr}(\cdot)$ and discriminator $D_{\rr}(\cdot)$, 
which generate RGB images from depth maps and distinguish generated RGB images from real RGB images, respectively.
$G_{\rr}(\cdot)$ employs the ResNet-18 architecture~\cite{he2016deep}, 
and $D_{\rr}(\cdot)$ follows the same architecture as $D(\cdot)$ except no condition inputs.

We adopt the objective functions of WGAN~\cite{arjovsky2017wasserstein} and CycleGAN~\cite{zhu2017unpaired} for training {\rdfcg}.
To be more specific, the \rdfcg loss includes
two discriminator losses ($\Loss_{D}$ and $\Loss_{D_{\rr}}$),
two generator losses ($\Loss_{G}$ and $\Loss_{G_{\rr}}$),
and a cycle loss ($\Loss_\text{cycle}$) as
\begin{align}
\label{equption:gan_dloss}
\Loss_{D} & =
  D \Large( G\left(\dr, \rr\right) | \rr \Large)
  - D (\dgt |\rr), \\
\Loss_{G} & =
  - D \Large( G\left(\dr, \rr\right) | \rr \Large), \\
\Loss_{D_\rr} & =
  D_\rr \Large( G_\rr\left(\dgt\right) \Large)
  - D_\rr (\rr), \\
\Loss_{G_\rr} & =
  - D_\rr \Large( G_\rr\left(\dgt\right) \Large), \\
\Loss_\text{cycle} & =
  {\LARGE\| G_\rr \Large( G \left(\dr, \rr\right) \Large) - \rr \LARGE\|}_1
  + {\LARGE\| G \Large( G_\rr \left(\dgt\right) \Large) - \dgt \LARGE\|}_1,
\end{align}
where the discriminator and generator losses only affect the corresponding discriminator and generator, respectively.

The overall loss for the {\rdfcg} branch $\Loss_\text{RDFC}$ combines all the loss terms above, i.e.,
\begin{align}
\Loss_\text{RDFC} = \Loss_{D} + \Loss_{G} + \Loss_{D_\rr} + \Loss_{G_\rr} + \Loss_\text{cycle}.
\end{align}

\subsection{\wada: Weighted Adaptive Instance Normalization}
\label{sec:connect_module}
To allow the depth feature information to guide the completion results of the RGB branch across all stages,
we design and apply the Weighted Adaptive Instance Normalization~(\wada) module,
which is first introduced in our preliminary work~\cite{wang2022rgb} and will be further elaborated below.

Inspired by StyleGAN~\cite{karras2019style} that uses AdaIN~\cite{huang2017arbitrary} to adapt a given style to the content and keeps the high-level content attributes,
the proposed {\wada} treats depth and RGB images as style and content inputs, respectively and mimics depth while keeping the semantic features of the RGB image.
By applying {\wada} at intermediate layers, the {\rdfcg} branch progressively absorbs the depth representations from the {\mcn} branch.

{\wada} is conducted via the following operations given an RGB image feature map $\fr \in \R^{h \times w \times C} $ from the {\rdfcg} branch's intermediate stage
and a depth feature map $\zz \in \R^{h \times w \times C}$ from {\mcn}'s corresponding stage,
where $h$, $w$, and $C$ are the height, width, and the number of channels (i.e., feature dimension), respectively.
For each channel $c$ ($1\le c \le C$), we compute the channel-wise scaled feature $y_s^{(c)}$ and bias $y_b^{(c)}$ as
\begin{align}
y_s^{(c)} & = \sigma(\zz^{(c)}) \frac{{\fr}^{(c)}-\mu(\fr^{(c)})}{\sigma(\fr^{(c)})} ,\\
y_b^{(c)} & = \mu(\zz^{(c)}),
\end{align}
where ${\fr}^{(c)}$ and ${\zz}^{(c)}$ are the $c$-th channel of the feature maps, and
$\mu(\cdot^{(c)})$ and $\sigma(\cdot^{(c)})$ are the spatial invariant channel-wise mean and variance, respectively.
Then, each element of \mbox{$\operatorname{\wada}(\zz,\fr) \in  \R^{h \times w \times C}$} is calculated by
\begin{equation}
\operatorname{\wada}(\zz,\fr)_{i,j,c} =
  y_s^{(c)} \operatorname{Attn}(\zz)_{i,j}
  +  y_b^{(c)} \operatorname{Attn}(\fr)_{i,j},
\label{equption:wada}
\end{equation}
where $1 \le i \le h$, $1 \le j \le w$, $1 \le c \le C$,
$\mathbf{x}_{i,j,c}$ and $\mathbf{x}_{i,j}$ respectively refers to the $(i,j,c)$-th and the $(i,j)$-th element of variable $\mathbf{x}$,
and $\operatorname{Attn}(\mathbf{x}) \in \R^{h \times w}$ is the self-attention result~\cite{vaswani2017attention} on the dimension reduction (by a $1 \times 1$ convolutional layer) of variable $\mathbf{x}  \in \R^{h \times w \times C}$.
Compared with the original AdaIN~\cite{huang2017arbitrary} with no learnable parameters, {\wada}, by self-attentions on both inputs, conduct more subtly control on the strength of either feature module in the fusion process, enhancing the overall coherence of the module’s output.

\subsection{Confidence Fusion Head}
\label{sec:confidence_fusion}

We follow our preliminary version~\cite{wang2022rgb} to combine the depth completion results from the two branches.
The local depth map $\dl$ generated by the {\mcn} branch relies more on valid raw depth information,
while the fused depth map $\df$ generated by the {\rdfcg} branch relies more on the textural RGB features.
In a confidence fusion head as shown in the right of Fig.~\ref{fig:architect}, we use the confidence maps~\cite{van2019sparse} to calculate the final depth prediction $\dpred$ by
\begin{equation}
\dpred(i, j)=\frac{e^{\cl(i, j)} \dl(i, j)+e^{\cf(i, j)} \df(i, j)}{e^{\cl(i, j)} + e^{\cf(i, j)}},
\end{equation}
where $1 \le i \le H$, $1 \le j \le W$, and $\mathbf{x}(i, j)$ refers to the $(i,j)$-th element of variable $\mathbf{x}$.
In the final depth prediction, the local and fused depth map contributes more to the accurate and noisy/missing regions of the raw depth map, respectively.
~\\

\begin{figure}[b]
\centering
\includegraphics[width=0.99\linewidth]{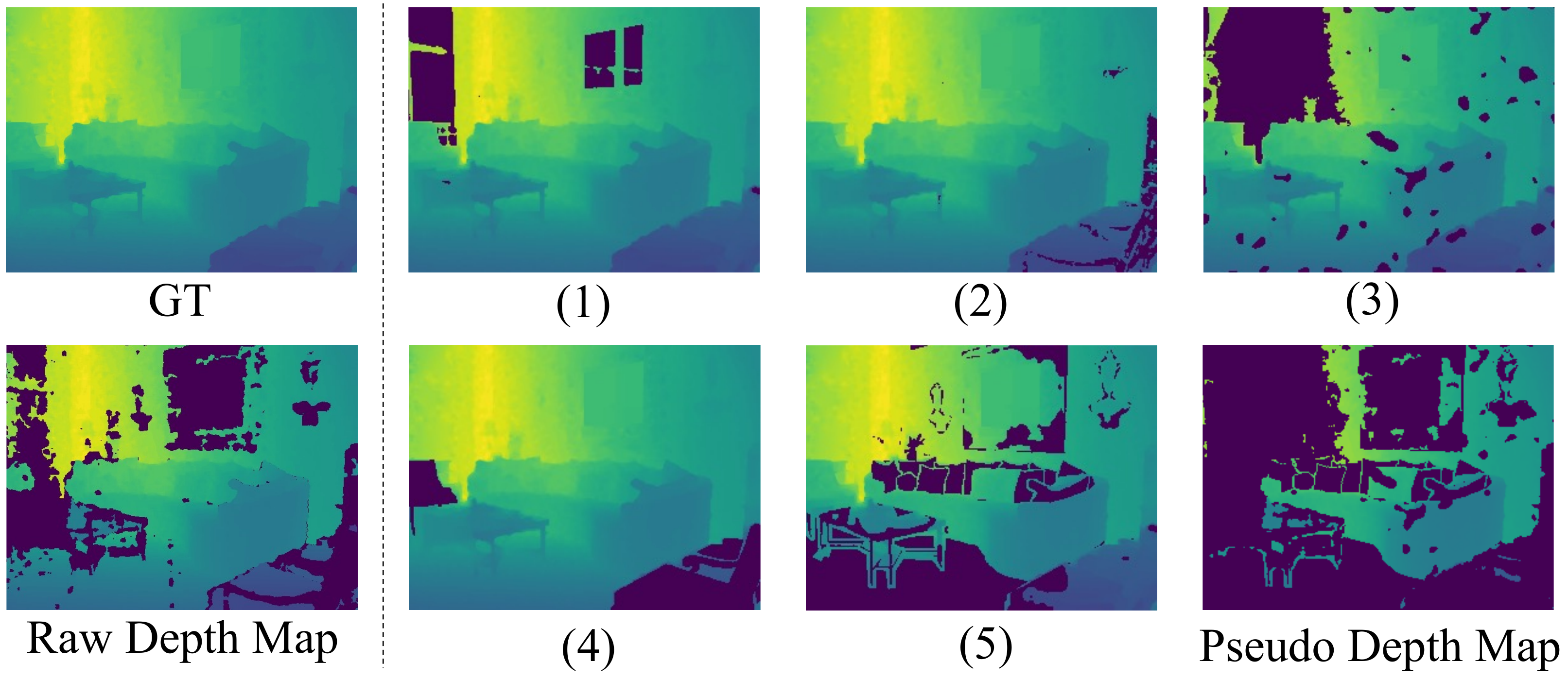}
\caption{Visualizations of the proposed pseudo depth map and five sampling methods.
`GT' refers to the reconstructed (ground-truth) depth map.
The shown pseudo depth map is generated from the raw depth map by applying all five sampling methods together.}
\label{fig:pseudo}
\end{figure} 
\subsection{Pseudo Depth Map for Training}
\label{sec:pseudo depth map}

To obtain a more robust depth completion model for \textit{indoor} scenarios, we adapt the pseudo depth map for training as in our preliminary work~\cite{wang2022rgb}.
As compared in Fig~\ref{fig:setting},
the commonly used random sparse sampling method~\cite{ma2018sparse,park2020non,lee2021depth} is not suitable for indoor scenarios for the significant differences of depth distributions and missing patterns.

The set of five proposed synthetic methods are as follows:

\begin{enumerate} [itemsep=1pt,topsep=1pt,parsep=1pt,leftmargin=20pt,label=(\arabic*)]
\item \textit{Highlight masking}.
  RGB-D cameras have difficulty in obtaining depth data of shiny surfaces because IR rays reflected from shiny surfaces are weak or scattered~\cite{kim2013depth},
  and smooth and shiny objects often lead to specular highlights and bright spots in the RGB images.
  Hence, we detect highlight regions in RGB images~\cite{arnold2010automatic} and mask them in depth maps to generate pseudo depth maps.

\item \textit{Black masking}.
  Dark and matte surfaces absorb rather than reflect radiations which strongly affected the depth map values~\cite{baek2020distance}.
  We randomly mask the depth pixels whose RGB values are all in $[0, 5]$ to directly mimic invalid depth values in dark regions.

\item \textit{Graph-based segmentation masking}.
  Chaotic light reflections in the complex environment interfere with the return of infrared light and cause discrete and irregular noises in depth maps.
  To simulate this phenomenon, we use graph-based segmentation~\cite{felzenszwalb2004efficient} to divide the RGB image into blocks and randomly mask some small blocks.

\item \textit{Semantic masking}.
  Some materials, such as glass, mirror, and porcelain surfaces, easily cause scattered infrared reflection and missing depth values.
  We utilize the semantic label information to locate objects probably containing these materials, such as televisions, mirrors, and windows, and we randomly mask one or two such objects (but keep depth pixels on their edges) in each frame.

\item \textit{Semantic XOR masking}.
  With the similar motivation to (3) graph-based segmentation masking, we use semantic segmentation to recognize complex regions and mask depth values in these regions.
  The complex region is defined as those whose predicted segmentation results, segmented by a U-Net~\cite{ronneberger2015u} trained on 20\% RGB images of the training set, are different from the ground truth. In other words, we conduct the Exclusive Or (XOR) operation on the segmentation results and the ground truth to obtain the masking.
\end{enumerate}

For each of the five methods, we independently randomly pick it with the probability of $50\%$, and we combine the masks from the picked methods to generate the final pseudo depth map from the raw depth. An example is shown in Fig~\ref{fig:pseudo}.

\subsection{Overall Loss Function}
\label{sec:loss_function}
We train all the network in an end-to-end way, with all previously described losses and the $L_1$ loss on the final prediction.
The overall loss function is defined as:
\begin{equation}
\Loss_\text{overall}= \Loss_\text{MCN} + \Loss_\text{RDFC} +  \lambda_\text{pred} {\| \dpred - \dgt \|}_1,
\end{equation}
where $\lambda_\text{pred}$ is the weight hyperparameter for the $L_1$ loss.

\section{Experiments}
\subsection{Datasets}
\label{sec:ex_datasets}
We conducted experiments on two widely-used benchmarks: \mbox{NYU-Depth V2}~\cite{Silberman:ECCV12} and \mbox{SUN RGB-D}~\cite{song2015sun}.

\para{NYU-Depth V2}
The NYU-Depth V2 dataset~\cite{Silberman:ECCV12} contains pairs of RGB and depth images collected from Microsoft Kinect in $464$ indoor scenes.
The dataset comprises densely labeled data samples divided into the training set with $795$ images and the test set with $654$ images.
Each sample includes an RGB image, a raw depth image captured by sensors, a reconstructed depth map treated as ground-truth labels, and a segmentation mask.
The dataset also has about $50,000$ unlabeled data samples with only RGB and raw images.
Following existing methods~\cite{ma2018sparse,park2020non}, we trained on the unlabeled images and the training set, and we used the test set for evaluation.
All images were resized to $320 \times 240$ and center-cropped to $304 \times 228$.

\para{SUN RGB-D}
The SUN RGB-D dataset~\cite{song2015sun} contains $10,335$ RGB-D images captured by four different sensors, offering a diverse and comprehensive collection of scenes that effectively facilitate the evaluation of model generalization.
Moreover, the dataset has dense semantic segmentation and 3D bounding box annotations that enables downstream task (e.g., object detection) evaluations.
Following the official dataset split~\cite{song2015sun}, we used $4,845$ images for training and $4,659$ for testing
and used the refined depth map derived from multiple frames~\cite{song2015sun} as the ground truths for evaluation.
All images were resized to $320 \times 240$ and randomly cropped to $304 \times 228$.

\subsection{Evaluation Metrics}
To comprehensively assess the performance of depth completion methods, we employed common metrics in both the original depth space and the point cloud space, as well as depth map and point cloud visualizations for qualitative evaluation.

\para{Depth Values}
We adopted three metrics that measure the depth values directly: the root mean squared error (\textit{RMSE}), the absolute relative error (\textit{Rel}), and $\delta_{th}$ as proposed by Ma~\etal~\cite{ma2018sparse}.

\textit{RMSE} is sensitive to substantial errors and offers valuable insight of the overall accuracy, which is defined as
\begin{equation}
	\text{RMSE} = \sqrt{\frac{1}{HW}\sum_{i, j}{\Large(\dpred(i,j) - \dgt(i, j)\Large)}^2}.
\end{equation}

\textit{Rel} assesses the relative error by normalizing the absolute deviation by the ground truth. \textit{Rel} is defined as
\begin{equation}
	\text{Rel} = \frac{1}{HW}\sum_{i, j}\frac{|\dpred(i, j) - \dgt(i, j)|}{\dgt(i, j)}.
\end{equation}

$\delta_{th}$ measures the percentage of predicted pixels whose relative error is within the relative threshold $th$.
The mathematical expression for $\delta_{th}$ is
\begin{equation}
	\delta_{th} = \frac{1}{HW}\sum_{i, j} \mathcal{I} \left( \max\left(\frac{\dpred(i, j)}{\dgt(i, j)}, \frac{\dgt(i, j)}{\dpred(i, j)}\right) <  th \right),
\end{equation}
where $\mathcal{I}(\cdot)$ is the indicator function.
With the same threshold value, a higher $\delta_{th}$ value indicates better consistency of the depth completion results.

\para{Point Clouds}
We noticed that the metrics on depth values effectively assess the global accuracy but inadequately address local outliers.
Consequently, we proposed to transform completed depth maps into point clouds and measured the Chamfer distance (\textit{CD}) and the averaged F1 score ($\textit{F}_{1}$) for a thorough evaluation.
Both \textit{CD} and $\textit{F}_{1}$ adeptly capture the geometric structure and relative positional relationships between point clouds, thereby exhibiting heightened sensitivity to local anomalies and noise.

To convert depth maps~($\dpred$ and $\dgt$) into point clouds~($\PPRED$ and $\PGT$),
we employed the following formula for each pixel $(i, j)$ in the depth map to get the corresponding point $\pp = (x, y, z)$ in the point cloud:
\begin{equation}
  {\left[x, y, z\right]}\T = \mathbf{d}\left(i, j\right) \mathbf{K}^{-1} {\left[i, j, 1\right]}\T,
\end{equation}
where $\mathbf{K}$ represents the intrinsic matrix of the camera.

The Chamfer distance (\textit{CD}) is a symmetric distance metric between two point clouds, defined as
\begin{align}
	\text{CD} =
    & \ \frac{1}{|\PGT|} \sum_{\pp \in \PGT} \min_{\pp' \in \PPRED} {\|\pp - \pp'\|}^2 \nonumber \\
	  & \ + \frac{1}{|\PPRED|} \sum_{\pp \in \PPRED} \min_{\pp' \in \PGT} {\|\pp - \pp'\|}^2,
\end{align}
where $|\PGT|$ and $|\PPRED|$ denotes the number of points in $\PGT$ and $\PPRED$, respectively,
$\pp$ and $\pp'$ denote points in the 3D space,
and $\|\cdot\|$ is the Euclidean distance.

The averaged F1 score ($\textit{F}_{1}$) is defined as the average of the
harmonic mean of precision ($\textit{Prec}_\Delta$) and recall ($\textit{Rec}_\Delta$) with a distance threshold $\Delta$ (Unit: meter):
\begin{align}
	\text{Prec}_\Delta & = \frac{1}{|\PPRED|} \sum_{\pp \in \PPRED} \mathcal{I} \left( \min_{\pp' \in \PGT} \|\pp - \pp'\| < \Delta \right), \\
	\text{Rec}_\Delta & = \frac{1}{|\PGT|} \sum_{\pp \in \PGT} \mathcal{I} \left( \min_{\pp' \in \PPRED} \|\pp - \pp'\| < \Delta \right), \\
  {\text{F}_1} & = \frac{1}{3} \sum_{\Delta \in \{0.02, 0.03, 0.04\}} \frac{2}{\text{Prec}_\Delta^{-1} + \text{Rec}_\Delta^{-1}},
\end{align}
where $\mathcal{I}(\cdot)$ is the indicator function and $\Delta$ determines whether two points are matched (i.e., closely enough).

\subsection{Implementation Details}
\label{sec:implement_details}

For the {\mcn} branch, the segmentation results were from a pre-trained and frozen PSPNet~\cite{zhao2017pyramid} with a ResNet-50 backbone,
and the normal map generator was a pre-trained U-Net~\cite{bae2021estimating} that was jointly trained with other modules.
The {\rdfcg} branch and other parts of the proposed network were trained from scratch.
The weights and bias in $G(\cdot)$, $G_{\rr}(\cdot)$, $D(\cdot)$, and $D_{\rr}(\cdot)$ were initialized from $\mathcal{N}(0, 0.02^2)$ and $0$, respectively.
The values of $\lambda_\text{l}$ and $\lambda_\text{pred}$ were set to $0.5$ and $5$, respectively.
The optimizer for {\mcn} was AdamW~\cite{loshchilov2018decoupled} with a weight decay of $0.01$ and an initial learning rate $lr_0$ of $0.002$.
The optimizers for other modules were Adam~\cite{kingma2014adam} with an initial learning rate $lr_0$ of $0.004$.
All optimizers had $\beta_1=0.5$, $\beta_2 = 0.999$.
We trained the network 150 epochs and used a linear learning rate scheduler for updates after the 100th epoch,
where the learning rate $lr_\text{epoch} = lr_0 \times \left(1 - \frac{\max(\text{epoch}, 100)-100}{50} \right)$.

\subsection{Training and Evaluation Settings}
To draw a comprehensive performance analysis, we set up three different evaluation schemes and their corresponding training strategies.
In the test phase, to predict and reconstruct depth maps (denoted as $\TT$), we used three different inputs in the three settings respectively,
which are the raw depth maps ($\RR$),
randomly-sampled sparse depth maps ($\RS$) from the raw depth maps,
and randomly-sampled sparse depth maps ($\TS$) from the reconstructed depth maps.
The three settings are as the following specified:
\begin{itemize}
  \item \textit{Setting A} ($\RR \Rightarrow \TT$):
    To be the most in line with the real scenario of indoor depth completion, we input a raw depth map without downsampling during testing.
    We used the pseudo depth maps as the input and supervised with the raw depth image, to train Sparse2Dense~\cite{ma2018sparse}, CSPN~\cite{cheng2018depth}, DeepLidar~\cite{qiu2019deeplidar}, NLSPN~\cite{park2020non}, GraphCSPN~\cite{liu2022graphcspn}, the preliminary model {\rdfg}~\cite{wang2022rgb}, and the proposed model {\rdfcg}.
    Meanwhile, we compared with DM-LRN~\cite{li2020multi} and MS-CHN~\cite{senushkin2020decoder} that were trained in the synthetic semi-dense sensor data~\cite{senushkin2020decoder}.

  \item \textit{Setting B} ($\RS \Rightarrow \TT$):
    Following a few works~\cite{ma2018sparse,cheng2018depth,qiu2019deeplidar,park2020non},
    we used a sparse depth map with 500 randomly sampled depth pixels of the \textit{raw} depth image as the input during testing.
    In the training stage, the input was the same as that for testing, but the ground truth was the raw depth due to the unavailability of completed depth maps.

  \item \textit{Setting C} ($\TS \Rightarrow \TT$):
    For comparing more existing methods~\cite{ma2018sparse,cheng2018depth,imran2019depth,lee2019depth,qiu2019deeplidar,park2020non,lee2021depth} that focus on depth completion in sparse scenes, 
    we used a sparse depth map with 500 randomly sampled depth pixels of the \textit{reconstructed} depth map as the input during testing.
    The training input and output ground truth were the same as those of \textit{Setting B}.
    As illustrated in Fig.~\ref{fig:setting}, the downsampled input in this setting reveals ground truth depth values that are unavailable in practice.

\end{itemize}

As we discussed, \textit{Setting A} ($\RR \Rightarrow \TT$) is the most plausible for indoor depth completion and is the primary focus of the problem we aim to address.
Therefore, \textit{Setting A} was used in all experiments.
We included the other two settings in the main experiments on {NYU-Depth V2} for comprehensive comparisons
and demonstrated the generality of our methods, i.e., the robustness and adaptability under different conditions.
~\\

\subsection{Comparisons with State-of-the-Art Methods}
\label{sec:compar_sota}
\begin{table}[h!]
  \caption{Quantitative results on the NYU-Depth V2 dataset.
  Spares2Dense and DGCG in {$\TS \Rightarrow \TT$} used 200 sampled pixels while others used 500 pixels.
  }
  \label{tab:nyu}
  \centering
  \resizebox{1\columnwidth}{!}{
    \begin{tabular}{c|c|c|c|ccc}
      \toprule
      Setting & Method & \textit{RMSE} $\downarrow$ & \textit{Rel} $\downarrow$
      & $\delta_{1.25} \uparrow$ & $\delta_{1.25^{2}} \uparrow$ & $\delta_{1.25^{3}} \uparrow$ \\
      \midrule
      \multirow{8.5}{*}{\makecell[cc]{\textit{Setting A}\\$\RR \Rightarrow \TT$}}
      & Sparse2Dense~\cite{ma2018sparse} & $0.538$ & $0.087$ & $87.1$ & $94.0$ & $97.0$ \\
      & CSPN~\cite{cheng2018depth} & $0.324$ & $0.051$ & $95.1$ & $98.4$ & $99.4$ \\
      & DeepLidar~\cite{qiu2019deeplidar} & $0.152$ & $0.016$ & $98.6$ & $99.7$ & $99.9$\\
      & MS-CHN~\cite{li2020multi} & $0.190$ & $0.018$ & $98.8$ & $99.7$ & $99.9$ \\
      & DM-LRN~\cite{senushkin2020decoder} & $0.205$ & $0.014$ & $98.8$ & $99.6$ & $99.9$ \\
      & NLSPN~\cite{park2020non} & $0.153$ & $0.015$ & $98.6$ & $99.6$ & $99.9$ \\
       & GraphCSPN~\cite{liu2022graphcspn} & $\underline{0.133}$ & $0.015$ & $98.7$ & $99.7$ & $99.9$ \\
      \cmidrule(lr){2-7}
      & {\rdfg}~\cite{wang2022rgb} & $0.139$ & $\underline{0.013}$ & $98.7$ & $99.6$ & $99.9$ \\
      & {\rdfcg} & $\mathbf{0.120}$ & $\mathbf{0.012}$ & $98.8$ & $99.7$ & $99.9$ \\
      \cmidrule{1-7}
      \multirow{7.5}{*}{\makecell[cc]{\textit{Setting B}\\$\RS \Rightarrow \TT$}}
      & Sparse2Dense~\cite{ma2018sparse} & $0.335$ & $0.060$ & $94.2$ & $97.1$ & $98.8$ \\
      & CSPN~\cite{cheng2018depth} & $0.500$ & $0.139$ & $85.7$ & $92.9$ & $96.3$ \\
      & DeepLidar~\cite{qiu2019deeplidar} & $0.288$ & $0.073$ & $93.6$ & $98.6$ & $99.7$\\
      & NLSPN~\cite{park2020non} & $0.348$ & $\mathbf{0.043}$ & $93.0$ & $96.7$ & $98.5$  \\
      & GraphCSPN~\cite{liu2022graphcspn} & $0.299$ & $0.082$ & $94.6$ & $98.7$ & $99.6$ \\
      & GAENet~\cite{chen2022depth} & $\underline{0.260}$ & $0.067$ & $94.7$ & $98.9$ & $99.7$ \\
      \cmidrule(lr){2-7}
      & {\rdfg}~\cite{wang2022rgb} & $0.309$ & $0.053$ & $93.6$ & $97.6$ & $99.0$ \\
      & {\rdfcg} & $\mathbf{0.242}$ & $\underline{0.047}$ & $96.1$ & $99.1$ & $99.7$ \\
      \cmidrule{1-7}
      \multirow{11.5}{*}{\makecell[cc]{\textit{Setting C}\\$\TS \Rightarrow \TT$}}
      & Sparse2Dense~\cite{ma2018sparse}  & $0.230$ & $0.044$ & $97.1$ & $99.4$ & $99.8$ \\
      & CSPN~\cite{cheng2018depth} & $0.117$ & $0.016$ & $99.2$ & $99.9$ & $100.0$\\
      & 3coeff~\cite{imran2019depth} & $0.131$ & $0.013$ & $97.9$ & $99.3$ & $99.8$ \\
      & DGCG~\cite{lee2019depth}  & $0.225$ & $0.046$ & $97.2$ & $-$ & $-$\\
      & DeepLidar~\cite{qiu2019deeplidar} & $0.115$ & $0.022$ & $99.3$ & $99.9$ & $100.0$ \\
      & NLSPN~\cite{park2020non} & $\underline{0.092}$ & $\mathbf{0.012}$ & $99.6$ & $99.9$ & $100.0$  \\
      & PRR~\cite{lee2021depth} & $0.104$ & $0.014$ & $99.4$ & $99.9$ & $100.0$\\
      & GraphCSPN~\cite{liu2022graphcspn} & $\mathbf{0.090}$ & $\mathbf{0.012}$ & $99.6$ & $99.9$ & $100.0$ \\
      & GAENet~\cite{chen2022depth} & $0.114$ & $0.018$ & $99.3$ & $99.9$ & $100.0$ \\
      \cmidrule(lr){2-7}
      & {\rdfg}~\cite{wang2022rgb} & $0.103$ & $0.016$ & $99.4$ & $99.9$ & $100.0$ \\
      & {\rdfcg} & $0.094$ & $\mathbf{0.012}$ & $99.6$ & $99.9$ & $100.0$ \\
      \bottomrule
    \end{tabular}
  }
\end{table} 
\begin{figure*}[t]
\centering
\includegraphics[width=.95\linewidth]{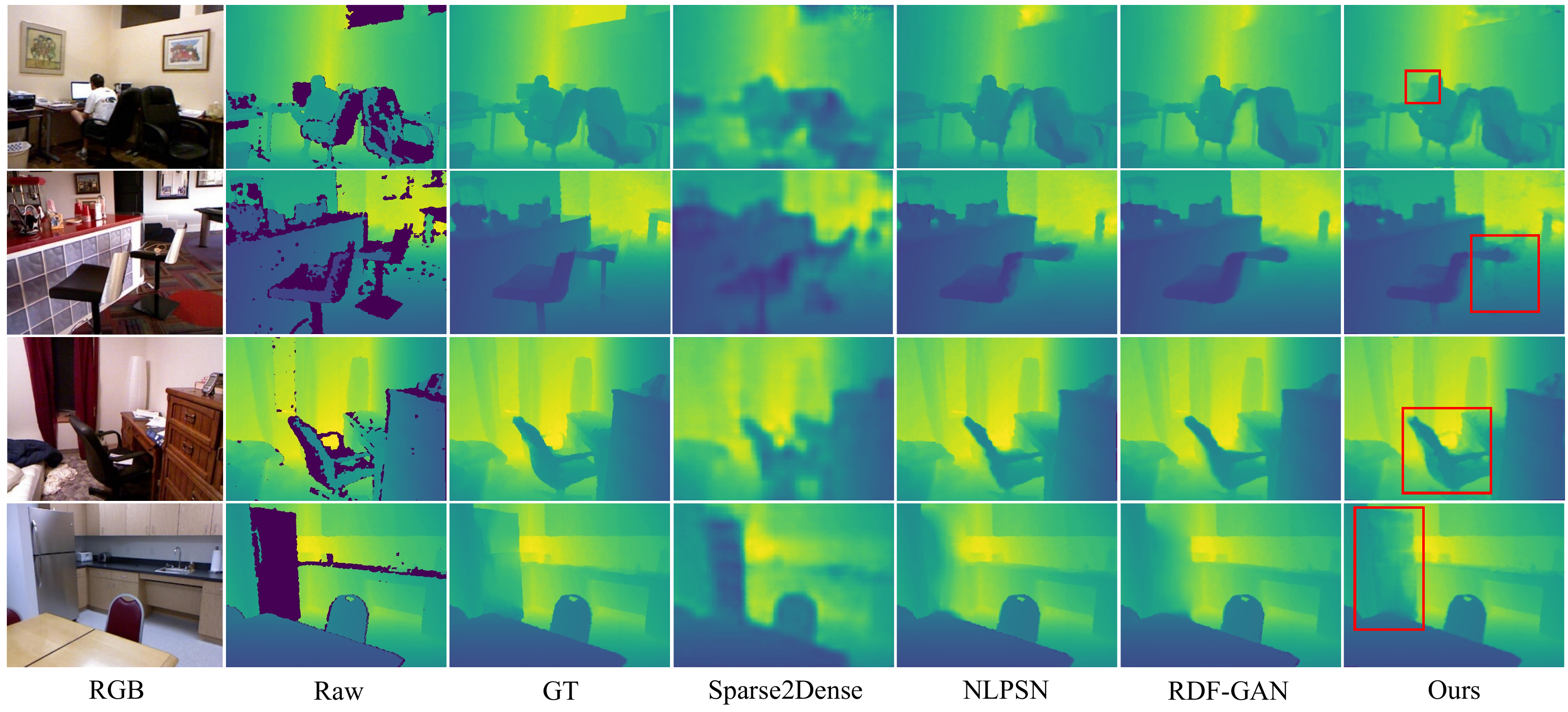}
\caption{Depth completion comparisons on NYU-Depth V2 with $\RR \Rightarrow \TT$.}
\label{fig:nyu_r2d}
\end{figure*}

\para{NYU-Depth V2}
The performance comparison on depth maps of our method and the other state-of-the-art methods on NYU-Depth V2 are shown in Tab.~\ref{tab:nyu}.
Given the results, we concluded the following:

\begin{itemize} [itemsep=1pt,topsep=1pt,parsep=1pt,leftmargin=10pt] 
  \item
    In the most realistic setting of $\RR \Rightarrow \TT$,
    compared to all the baselines, {\rdfcg} had significantly superior performance and obtained moderate improvement over the previous {\rdfg},
    leading to remarkably \textit{RMSE} of $0.120$ and \textit{Rel} of $0.012$.

  \item
    We selected a few representative scenes and visualized the completion results from different methods with the setting of $\RR \Rightarrow \TT$ in Fig.~\ref{fig:nyu_r2d}.
    {\rdfcg} produced more accurate and textured depth predictions in the missing depth regions. For example, the results within the red boxes clearly depicted the contour and depth information of subtle objects (laptops and chairs) and large missing ones (doors).

  \item In the setting of $\RS \Rightarrow \TT$,
    {\rdfcg} outperformed the baselines with big margins on \textit{RMSE},
    and achieved the best on all $\delta_{th}$ metrics and the second best on \textit{Rel}.
    Also, {\rdfcg} improved {\rdfg} substantially by a $22\%$ relative improvement on \textit{RMSE},
    indicating the efficacy of the newly proposed CycleGAN and Manhattan constraint components.

  \item We observed a similar trend in the setting of $\TS \Rightarrow \TT$ that {\rdfcg} obtained the best on four of all five metrics.
    In terms of \textit{RMSE}, {\rdfcg} without any iteration processing was only lower than NLSPN~\cite{qiu2019deeplidar} and GraphCSPN~\cite{liu2022graphcspn}  (but $1.2\times$ and {$1.5\times$} faster in inference than them, respectively).
    The results are commendable because {\rdfcg} is not designed for the sparse setting.
\end{itemize}

\begin{table}[h!]
  \caption{Quantitative results on {SUN RGB-D} in \textit{Setting A} ($\RR \Rightarrow \TT$).}
  \label{tab:sun-rgbd}
  \centering
  \resizebox{0.95\columnwidth}{!}
  {
    \begin{tabular}{c|c|c|ccc}
    \toprule
    Method & \textit{RMSE} $\downarrow$ & \textit{Rel} $\downarrow$
    & $\delta_{1.25} \uparrow$ & $\delta_{1.25^{2}} \uparrow$ & $\delta_{1.25^{3}} \uparrow$ \\
    \midrule
    Sparse2Dense~\cite{ma2018sparse} & $0.329$ & $0.074$ & $93.9$ & $97.0$ & $98.1$\\
    CSPN~\cite{cheng2018depth} & $0.295$ & $0.137$ & $95.6$ & $97.5$ & $98.4$ \\
    DeepLidar~\cite{qiu2019deeplidar} & $0.279$ & $0.061$ & $96.9$ & $98.0$ & $98.4$\\
    MS-CHN~\cite{li2020multi} & $0.235$ & $0.046$ & $96.1$ & $98.6$ & $99.4$ \\
    DM-LRN~\cite{senushkin2020decoder} & $0.268$ & $0.069$ & $95.3$ & $98.0$ & $99.1$ \\
    NLSPN~\cite{park2020non} & $0.267$ & $0.063$ & $97.3$ & $98.1$ & $98.5$ \\
    GraphCSPN~\cite{liu2022graphcspn} & $\underline{0.232}$ & $\underline{0.049}$ & $96.9$ & $98.4$ & $99.0$ \\
    \cmidrule(lr){1-6}
    {\rdfg}~\cite{wang2022rgb} & $0.255$ & $0.059$ & $96.9$ & $98.4$ & $99.0$ \\
    {\rdfcg} & $\mathbf{0.214}$ & $\mathbf{0.040}$ & $97.0$ & $99.1$ & $99.8$ \\
    \bottomrule
    \end{tabular}
  }
\end{table} 

\para{SUN RGB-D}
The results on {SUN RGB-D} in \textit{Setting A} are shown in Tab.~\ref{tab:sun-rgbd}.
We observed the following:
\begin{itemize} [itemsep=1pt,topsep=1pt,parsep=1pt,leftmargin=10pt] 
  \item
    The depth completion task on {SUN RGB-D} is much more difficult than that on {NYU-Depth V2}.
    This may be due to the fact that {SUN RGB-D} encompasses a greater variety of scenes is sourced from various sensors.
    Nevertheless, {\rdfcg} achieved the best performance in all metrics (e.g., $0.214$ v.s. $0.232$ on \textit{RMSE} and $0.040$ v.s. $0.049$ on \textit{Rel}, compared with the second best method).

  \item
    As the threshold value of $\delta_{th}$ increased, the performance gap between {\rdfcg} and the best baseline enlarged (from $-0.3$ of $\delta_{1.25}$ to $+1.2$ of $\delta_{{1.25}^3}$).
    The results indicate that baselines failed to complete depth in some regions even the tolerance threshold went larger, while {\rdfcg} was more robust to local outliers.

  \item From the visualization results in Fig.~\ref{fig:sunrgb}, {\rdfcg} complemented the missing depth with detailed texture information for all different sensors, showing its great generality.
\end{itemize}

\begin{figure*}[t]
  \centering
  \includegraphics[width=0.95\linewidth]{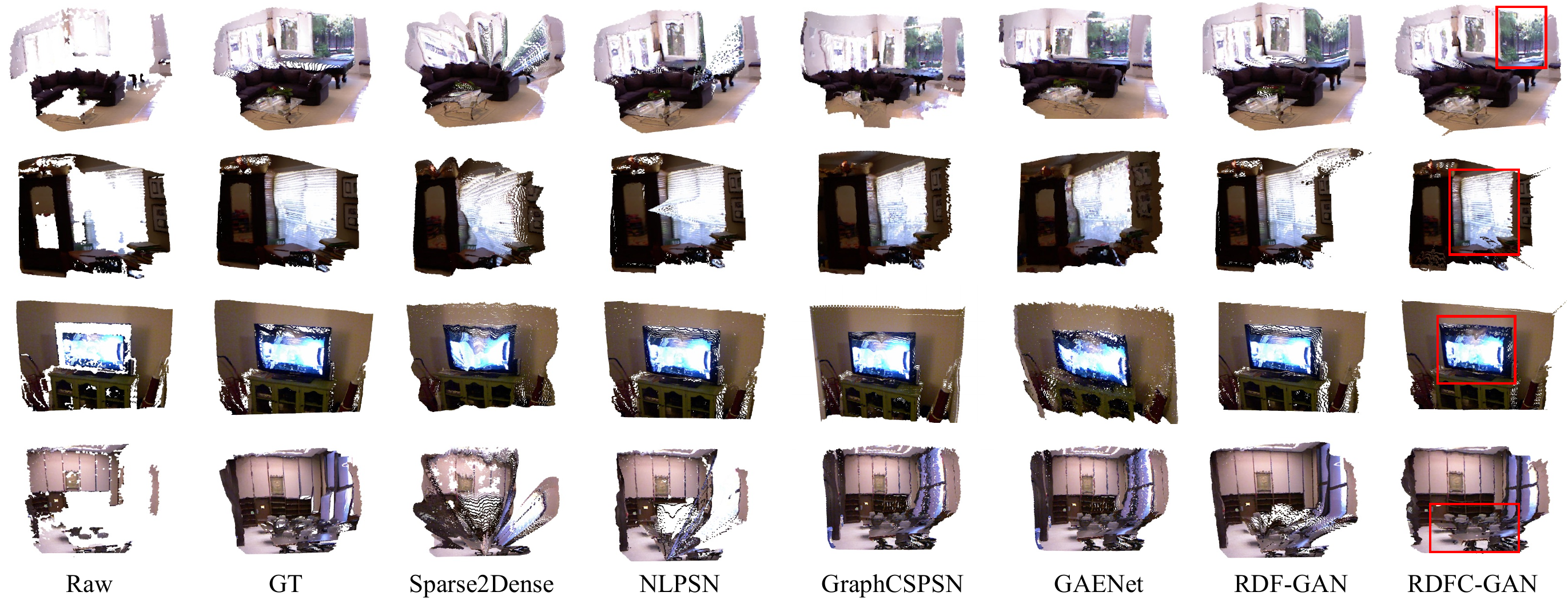}
  \caption{
  Depth completion comparisons by point cloud visualizations on {NYU-Depth V2} with $\RR \Rightarrow \TT$.
  }
  \label{fig:nyu_pc}
\end{figure*} 
\begin{table}[h!]
  \caption{
    Quantitative comparisons in point clouds on {NYU-Depth V2}.
    Unit of \textit{CD}: $1 \times 10^{-4}$ meter.
  }
  \label{tab:nyu-pc}
  \centering
  \resizebox{0.95\columnwidth}{!}{
    \begin{tabular}{c|cc|cc|cc}
      \toprule
      \multirow{2.5}{*}{Method}
      & \multicolumn{2}{c|}{\makecell[cc]{\textit{Setting A}\\$\RR \Rightarrow \TT$}}
      & \multicolumn{2}{c|}{\makecell[cc]{\textit{Setting B}\\$\RS \Rightarrow \TT$}}
      & \multicolumn{2}{c}{\makecell[cc]{\textit{Setting C}\\$\TS \Rightarrow \TT$}} \\
      \cmidrule(lr){2-3} \cmidrule(lr){4-5} \cmidrule(lr){6-7}
      & \textit{CD} & $\textit{F}_{1}$ & \textit{CD} & $\textit{F}_{1}$ & \textit{CD} & $\textit{F}_{1}$ \\
      \midrule
      Sparse2Dense~\cite{ma2018sparse} & $526.26$ & $0.73$ & $531.59$ & $0.69$ & $808.55$ & $0.58$ \\
      CSPN~\cite{cheng2018depth} & $42.35$ & $0.86$ & $334.51$ & $0.82$ & $174.33$ & $0.88$ \\
      NLSPN~\cite{park2020non} & $\underline{35.83}$ & $0.88$ & $342.99$ & $0.83$ & $92.01$ & $0.89$  \\
      GraphCSPN~\cite{liu2022graphcspn} & $52.28$ & $\underline{0.94}$ & $296.88$ & $\underline{0.86}$ & $89.40$ & $0.90$  \\
      GAENet~\cite{chen2022depth} & $710.36$ & $0.89$ & $\underline{267.07}$ & $\underline{0.86}$ & $\mathbf{47.39}$ & $\mathbf{0.95}$  \\
      \cmidrule(lr){1-7}
      \rdfg~\cite{wang2022rgb} & $79.66$ & $0.90$ & $284.10$ & $\underline{0.86}$ & $90.03$ & $0.93$ \\
      \rdfcg & $\mathbf{33.15}$ & $\mathbf{0.95}$ & $\mathbf{249.05}$ & $\mathbf{0.87}$ & $\underline{79.74}$ & $\underline{0.94}$ \\
      \bottomrule
    \end{tabular}
  }
\end{table} 
\para{Comparisons on Point Clouds}
In order to examine local accuracies and provide comprehensive comparisons, we selected a few representative baselines, converted their output depth maps on {NYU-Depth V2} to point clouds, and measured performance with the point clouds.
Based on the quantitative results in Tab.~\ref{tab:nyu-pc} and the visualization results in Fig.~\ref{fig:nyu_pc}, we can draw the following conclusions:
\begin{itemize}
  \item {\rdfcg} obtained the lowest Chamfer distance values and the highest averaged F1 scores in the first two settings and the second best in the other setting,
      indicating the superior performance in various experimental settings, especially the addressed indoor scenario.

  \item The visualization clearly demonstrated that {\rdfcg} completed depth maps of the missing regions stable and reasonable, while other methods made distorted or even incomplete estimations. The results highlighted the effectiveness of our proposed method in achieving more accurate completion results.
\end{itemize}

\subsection{Ablation Studies}
\label{sec:ablation}
We conducted ablation studies on \mbox{NYU-Depth V2} with the setting of $\RR \Rightarrow \TT$ that best reflects indoor scenarios.

\begin{table}[h!]
  \caption{
    Ablation study results for the Manhattan normal module.
    `PT' refers to pre-training on ScanNet.
    `FT' refers to pre-training on ScanNet and fine-tuning on NYU-Depth V2.
  }
  \label{tab:ablation-mcn}
  \centering
  \resizebox{0.99\columnwidth}{!}{
    \begin{tabular}{c|c|ccc}
      \toprule
      Case \# & {\mcn} Structures & \textit{RMSE} $\downarrow$ & \textit{Rel} $\downarrow$ & $\delta_{1.25}$ $\uparrow $ \\
      \midrule
      A-1 & Local Guidance~\cite{wang2022rgb} & $0.146$ & $0.021$ & $98.633$\\
      A-2 & Normal Generator (PT) & $0.147$ & $0.020$ & $98.584$ \\
      A-3 & Normal Generator (FT) & $0.132$ & $0.020$ & $98.712$ \\
      \cmidrule(lr){1-5}
      A-4 & Manhattan Normal Module (MNM) & $\mathbf{0.120}$ & $\mathbf{0.012}$ & $\mathbf{98.848}$ \\
      \cmidrule(lr){1-5}
      A-5 & MNM + Segmentation Features & $0.122$ & $0.013$ & $98.819$ \\
      \bottomrule
    \end{tabular}
  }
\end{table}

\para{The {\mcn} Branch}
\paragraph{Branch Structure}
We evaluated the proposed {\mcn} structure with the best alternative (i.e., the Local Guidance module) from our earlier model {\rdfg}~\cite{wang2022rgb} with the performance comparison shown in Tab.~\ref{tab:ablation-mcn}.
With only the pre-trained normal generator (Case A-2), the model performed comparably with the local guidance (Case A-1),
which may be due to their similar network structure (U-Net).
The fine-tuning step (Case A-3) enhanced the capacity of the normal generator with the \textit{RMSE} boosted from $0.147$ to $0.132$.
Using the segmentation network and the corresponding losses (Case A-4) further improved the performance.
We also included the features from the segmentation network as extra inputs to the normal generator (Case A-5), but its performance is slightly worse. We argue that the normal generator only needs to identify normals for different parts instead of utilizing the semantic features.

\begin{table}[h!]
  \caption{
    Ablation study results for the loss used in Manhattan normal module.
  }
  \label{tab:ablation-mcn-loss}
  \centering
  \resizebox{0.99\columnwidth}{!}{
    \begin{tabular}{c|c|ccc}
      \toprule
      Case \# & {\mcn} Loss Terms & \textit{RMSE} $\downarrow$ & \textit{Rel} $\downarrow$ & $\delta_{1.25}$ $\uparrow $\\
      \midrule
      B-1 & $\Loss_\vn$ only & $0.132$ & $0.020$ & $98.693$ \\
      B-2 & $\Loss_\vn + \Loss_\text{floor}$ & $0.127$ & $0.019$ & $98.787$ \\
      B-3 & $\Loss_\vn + \Loss_\text{ceiling}$ & $0.130$ & $0.019$ & $98.732$ \\
      B-4 & $\Loss_\vn + \Loss_\text{wall}$ & $0.123$ & $0.014$ & $98.806$ \\
      \cmidrule(lr){1-5}
      B-5 & $\Loss_\vn + \Loss_\text{floor} + \Loss_\text{ceiling} + \Loss_\text{wall}$ & $0.120$ & $\mathbf{0.012}$ & ${98.848}$ \\
      \cmidrule(lr){1-5}
      B-6 & $\Loss_\vn + \Loss_\text{MWA}$ & $\mathbf{0.117}$ & $0.013$ & $\mathbf{98.913}$ \\
      \bottomrule
    \end{tabular}
  }
\end{table} 
\paragraph{Branch Loss}
We conducted ablation studies for the losses introduced in the MCN branch.
As shown in Tab.~\ref{tab:ablation-mcn-loss}, each loss term plays a vital role in achieving accurate normal estimation~(Cases B-2 to B-4), and combining them together~(Case B-5) is better than using each of them.
Among the three losses, $\Loss_\text{wall}$ contributes the most.
We also compared a loss that directly modeling the normal orthogonality and parallel (Case B-6) as follows:
\begin{equation}
\Loss_\text{WMA} =
  \frac{\sum_{p\in \mathcal{P}_\text{w}, p' \in \mathcal{P}_\text{f} \cup \mathcal{P}_\text{c}} \frac{|\vn_p \cdot \vn_{p'}|}{\|\vn_p\| \|\vn_{p'}\|}}
    {\left|\mathcal{P}_\text{w}\right| \left( \left|\mathcal{P}_\text{f}\right| + \left|\mathcal{P}_\text{c}\right| \right)} +
  \frac{\sum_{p\in \mathcal{P}_\text{f}, p' \in \mathcal{P}_\text{c}} \frac{\vn_p \cdot \vn_{p'}}{\|\vn_p\| \|\vn_{p'}\|}}
    {\left|\mathcal{P}_\text{f}\right| \left|\mathcal{P}_\text{c}\right| },
\end{equation}
where $\mathcal{P}_\text{w}$, $\mathcal{P}_\text{f}$, and $\mathcal{P}_\text{c}$ are set of points of wall, floor, and ceiling, respectively.
$\Loss_\text{WMA}$ achieved comparable performance with a higher complexity (i.e., $\mathcal{O}(n^2)$ for $n$ points) but can be used where the camera exhibits roll and pitch rotations.

\begin{table}[h!]
  \caption{
    Ablation study results for the CycleGAN.
    For Cases B-1 and B-2, an $L_1$ loss on the fusion depth map is added~\cite{wang2022rgb}.
  }
  \label{tab:ablation-cyclegan}
  \centering
  \resizebox{0.85\columnwidth}{!}{
    \begin{tabular}{c|c|ccc}
      \toprule
      Case \# & GAN Structures & \textit{RMSE} $\downarrow$ & \textit{Rel} $\downarrow$ & $\delta_{1.25}$ $\uparrow $ \\
      \midrule
      C-1 & No GAN & $0.176$ & $0.031$ & $98.192$ \\
      C-2 & GAN & $0.129$ & $0.017$ & $98.818$ \\
      \cmidrule(lr){1-5}
      C-3 & CycleGAN & $\mathbf{0.120}$ & $\mathbf{0.012}$ & $\mathbf{98.848}$ \\
      \bottomrule
    \end{tabular}
  }
\end{table} 
\para{The {\rdfcg} Branch}
Tab.~\ref{tab:ablation-cyclegan} shows the ablation study results for the GAN branch.
The model without GAN (Case C-1) degenerated to a dual encoder-decoder structure.
In that case, the completed depth maps were shaped towards a blurry one and the results were poor.
Adding the GAN structure (Case C-2) substantially improved the performance,
and using the CycleGAN~\cite{CycleGAN2017} structure (Case C-3) led to further improvement (from $0.129$ to $0.120$ in terms of \textit{RMSE}).

\begin{table}[h!]
  \caption{
    Ablation study results for the {\wada} module.
  }
  \label{tab:ablation-wada}
  \centering
  \resizebox{0.9\columnwidth}{!}{
    \begin{tabular}{c|c|ccc}
      \toprule
      Case \# & Fusion Modules & \textit{RMSE} $\downarrow$ & \textit{Rel} $\downarrow$ & $\delta_{1.25}$ $\uparrow $ \\
      \midrule
      D-1 & IN~\cite{ulyanov2017improved} & $0.126$ & $0.016$ &  $98.817$ \\
      D-2 & AdaIN~\cite{huang2017arbitrary} & $0.133$ & $0.025$ &  $98.761$ \\
      \cmidrule(lr){1-5}
      D-3 & {\wada} & $\mathbf{0.120}$ & $\mathbf{0.012}$ & $\mathbf{98.848}$ \\
      \bottomrule
    \end{tabular}
  }
\end{table} 
\para{The {\wada} Module}
As shown in Tab.~\ref{tab:ablation-wada},
for the multi-stage fusion modules, {\wada} (Case D-1) outperformed the alternatives, i.e., instance normalization~(IN)~\cite{ulyanov2017improved} (Case D-2) and AdaIN~\cite{huang2017arbitrary} (Case D-3), by a clear margin.
We also observed similar trend as in {\rdfg}~\cite{wang2022rgb} that AdaIN was slightly inferior to the original IN,
indicating that directly applying the adaptive method may not for depth completion and our attention-based {\wada} is essential.

\begin{figure}[h!]
\centering
\includegraphics[width=.95\linewidth]{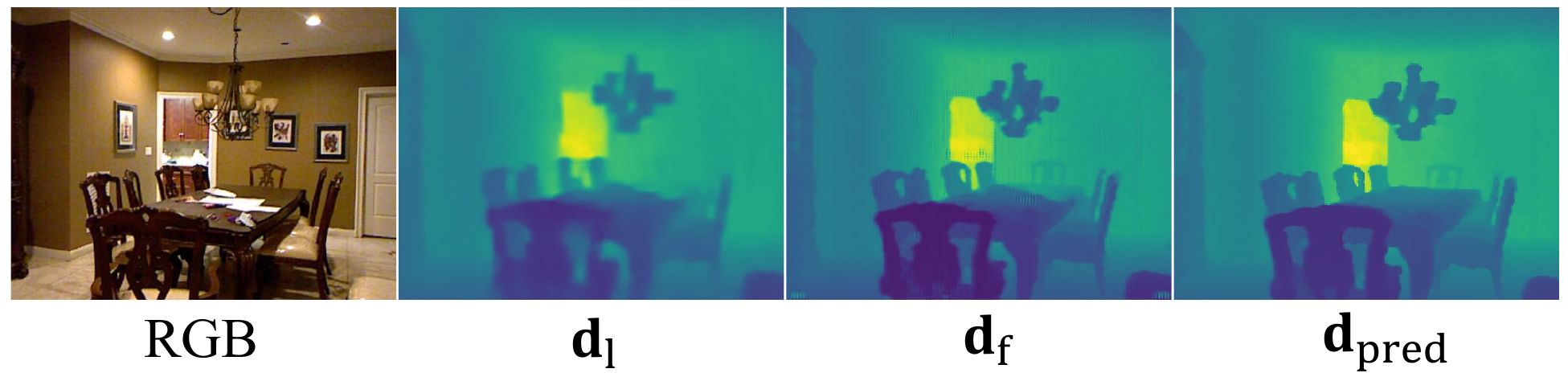}
\caption{Depth result visualizations of the {\mcn} branch ($\dl$), the {\rdfcg} branch ($\df$), and the entire model ($\dpred$).
}
\label{fig:ablation-branch}
\end{figure} 
\para{The Two-Branch Structure}
In Fig.~\ref{fig:ablation-branch}, we provide visualizations of three depth map completion outputs: $\dl$ from the {\mcn} branch, $\df$ from the {\rdfcg} branch, and the final output $\dpred$.
The {\mcn} branch generated a precise depth map, albeit lacking distinct contours.
The {\rdfcg} branch generated a depth map with more detailed textures while introducing a few noises and outliers.
Significantly, with the help of the confidence fusion head, the complete {\rdfcg} model produced a final completion that is both precise and robust, taking advantage of both two branches.

\begin{table}[h!]
    \centering
    \caption{
    Comparisons of 3D object detection results with the completed depth map on {SUN RGB-D}.
    The last column is the depth completion performance.
    }
    \resizebox{0.95\columnwidth}{!}{
    \begin{tabular}{c|cc|c}
    \toprule
         Method & \textit{mAP@25} $\uparrow$ & \textit{mAP@50} $\uparrow$ & \textit{RMSE} $\downarrow$ \\
         \midrule
         VoteNet~\cite{qi2019deep} & $59.07$ & $35.77$ & -- \\
         DeepLidar~\cite{qiu2019deeplidar} + VoteNet~\cite{qi2019deep} & $59.73$ & $35.49$ & $0.279$ \\
         NLSPN~\cite{park2020non} + VoteNet~\cite{qi2019deep} & $47.43$ & $26.10$ & $0.267$ \\
         {\rdfg}~\cite{wang2022rgb} + VoteNet~\cite{qi2019deep} & $\underline{60.64}$ & $\underline{37.28}$ & $\underline{0.255}$ \\
         {\rdfcg} + VoteNet~\cite{qi2019deep} & $\mathbf{61.02}$ & $\mathbf{37.47}$ & $\mathbf{0.214}$ \\
         GT + VoteNet~\cite{qi2019deep} & $59.44$ & $36.30$ & -- \\
         \cmidrule(lr){1-4}
         H3DNet~\cite{zhang2020h3dnet} & $60.11$ & $39.04$ & -- \\
         DeepLidar~\cite{qiu2019deeplidar} + H3DNet~\cite{zhang2020h3dnet} & $60.35$ & $39.16$ & $0.279$ \\
         NLSPN~\cite{park2020non} + H3DNet~\cite{zhang2020h3dnet} & $27.10$ & $9.77$ & $0.267$ \\
         {\rdfg}~\cite{wang2022rgb} + H3DNet~\cite{zhang2020h3dnet} & $\underline{61.03}$ & $\underline{39.71}$ & $\underline{0.255}$ \\
         {\rdfcg} + H3DNet~\cite{zhang2020h3dnet} & $\mathbf{61.75}$ & $\mathbf{40.63}$ & $\mathbf{0.214}$ \\
         GT + H3DNet~\cite{zhang2020h3dnet} & $60.51$ & $39.22$ & -- \\
    \bottomrule
    \end{tabular}
    }
    \label{tab:3ddet}
\end{table}

\subsection{Object Detection on the Completed Depth Map}
\label{sec:3ddet}
We conducted extended experiments using completed depth maps as the input for 3D object detection on \mbox{SUN RGB-D} to evaluate the quality of our depth completions.
Two SOTA models, VoteNet~\cite{qi2019deep} and H3DNet~\cite{zhang2020h3dnet}, were used as the detectors.
Tab.~\ref{tab:3ddet} shows that both detectors obtained a moderate improvement with our completed depth map.
Meanwhile, DeepLidar~\cite{qiu2019deeplidar} improved little in terms of the detection metrics;
NLSPN~\cite{park2020non} produced too much noise in the completion and even impaired the detection performance.
Using the ground-truth depth maps (provided by SUN-RGBD) as the input outperformed all others except for {\rdfg} and {\rdfcg}.
The reason is that the ground truth in SUN-RGBD is calculated by integrating multiple frames and still suffers missing depth areas, leading to suboptimal detection performance.
The results not only highlight the superiority of our approach but also showcase its robustness.

\section{Conclusion}
In this work, we propose a novel two-branch end-to-end network, {\rdfcg}, for indoor depth completion.
We design an RGB-depth fusion CycleGAN model to produce the fine-grained textural depth map and restrain it by a \mbox{Manhattan-constraint} network.
In addition, we propose a novel and effective sampling method to produce pseudo depth maps for training indoor depth completion models.
Extensive experiments have demonstrated that our proposed solution achieves state-of-the-art on the \mbox{NYU-Depth V2} and \mbox{SUN RGB-D} datasets.

\section*{Acknowledgments}
This work was supported in part by the National Key R\&D Program of China under Grant 2022YFF0904304,
the National Natural Science Foundation of China under Grant 62202065,
Shanghai Pujiang Program under Grant 21PJ1420300,
the Science and Technology Innovation Action Plan of Shanghai under Grant 22511105400,
and the BUPT Excellent Ph.D. Students Foundation under Grant CX2022224.


\bibliographystyle{IEEEtran}
\bibliography{reference}

\begin{IEEEbiography}[{\includegraphics[width=1in,height=1.25in, clip, keepaspectratio]{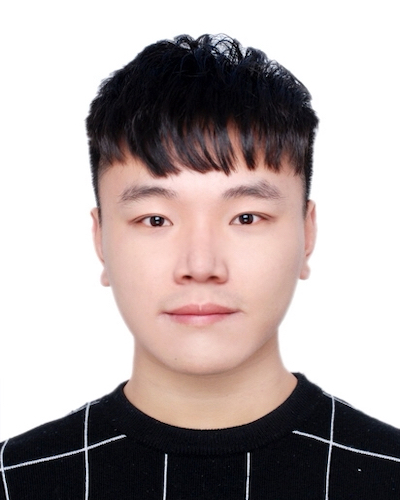}}]{Haowen Wang}
(Graduate Student Member, IEEE) is currently working towards a Ph.D. degree at the State Key Laboratory of Networking and Switching Technology, Beijing University of Posts and Telecommunications, Beijing, China. His current research interests include Deep Learning and Computer Vision, especially 3D perception, 3D reconstruction, and scene understanding.
\end{IEEEbiography}

\begin{IEEEbiography}[{\includegraphics[width=1in,height=1.25in, clip, keepaspectratio]{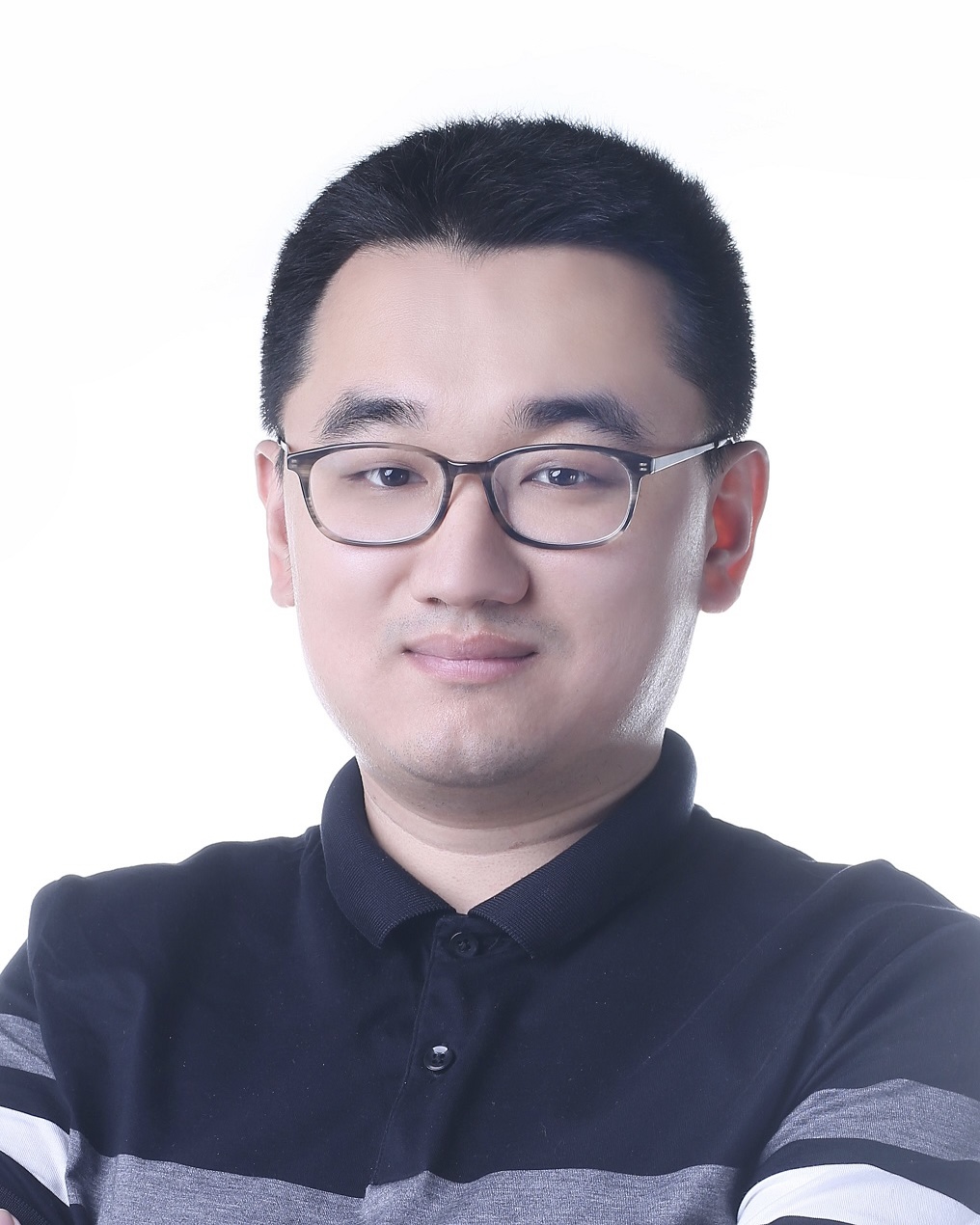}}]{Zhengping Che}
(Member, IEEE) received the Ph.D. degree in Computer Science from the University of Southern California, Los Angeles, CA, USA, in 2018,
and the B.E. degree in Computer Science (Yao Class) from Tsinghua University, Beijing, China, in 2013.
He is now with Midea Group.
His research interests lie in the areas of deep learning, embodied AI, computer vision, and temporal data analysis.
\end{IEEEbiography}

\begin{IEEEbiography}[{\includegraphics[width=1in,height=1.25in, clip, keepaspectratio]{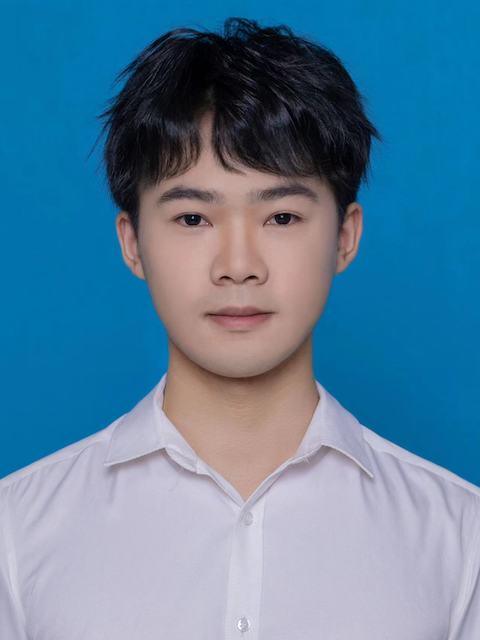}}]{Yufan Yang} is currently working towards the M.Eng. degree at the State Key Laboratory of Networking and Switching Technology, Beijing University of Posts and Telecommunications, Beijing, China. His research interests include computer vision, and deep learning.
\end{IEEEbiography}

\begin{IEEEbiography}[{\includegraphics[width=1in,height=1.25in, clip, keepaspectratio]{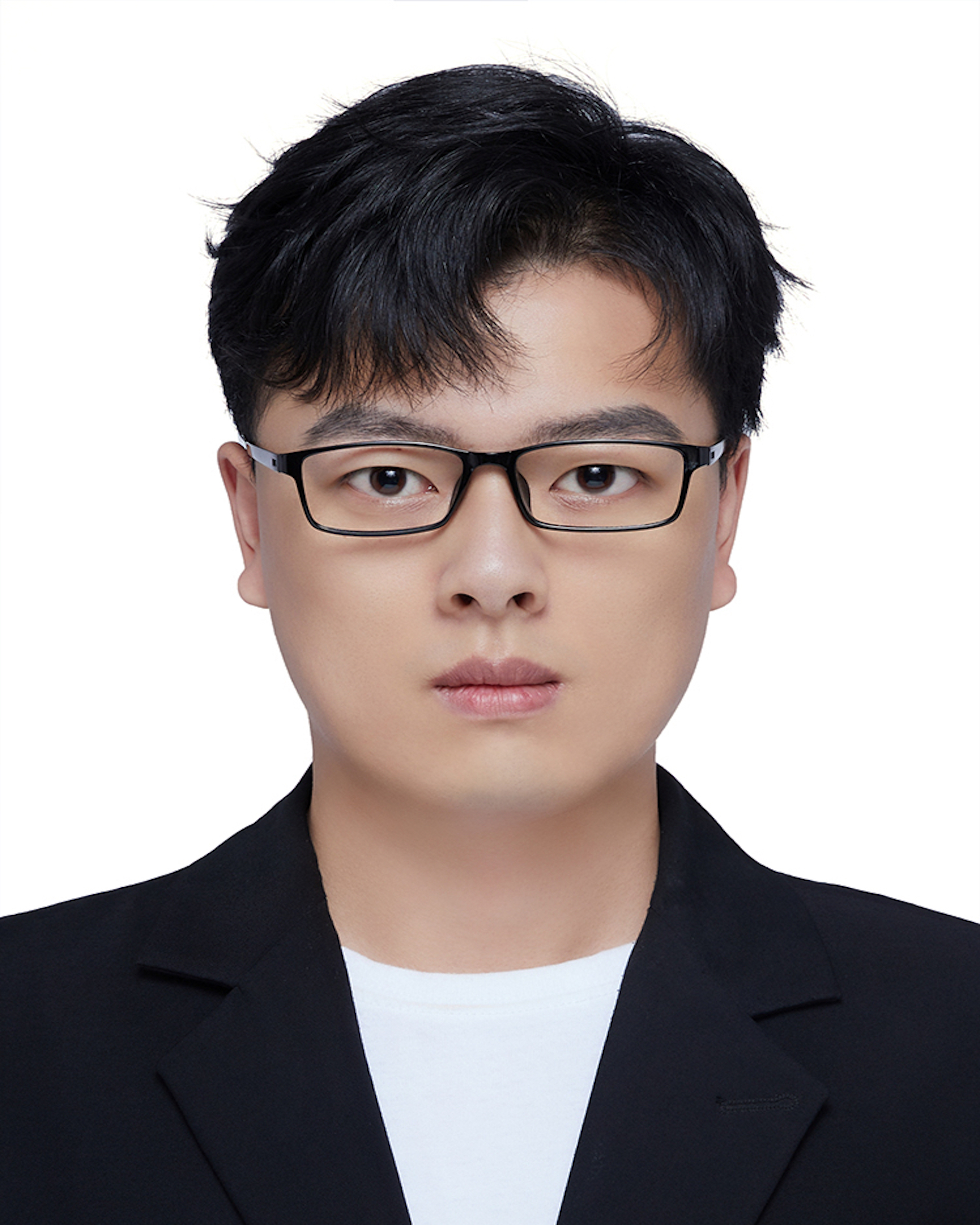}}]{Mingyuan Wang}
is currently a master student at the State Key Laboratory of Network and Switching Technology,
Beijing University of Posts and Telecommunications, Beijing, China.
His research focuses on Computer Vision and 3D reconstruction.
\end{IEEEbiography}

\begin{IEEEbiography}[{\includegraphics[width=1in,height=1.25in, clip, keepaspectratio]{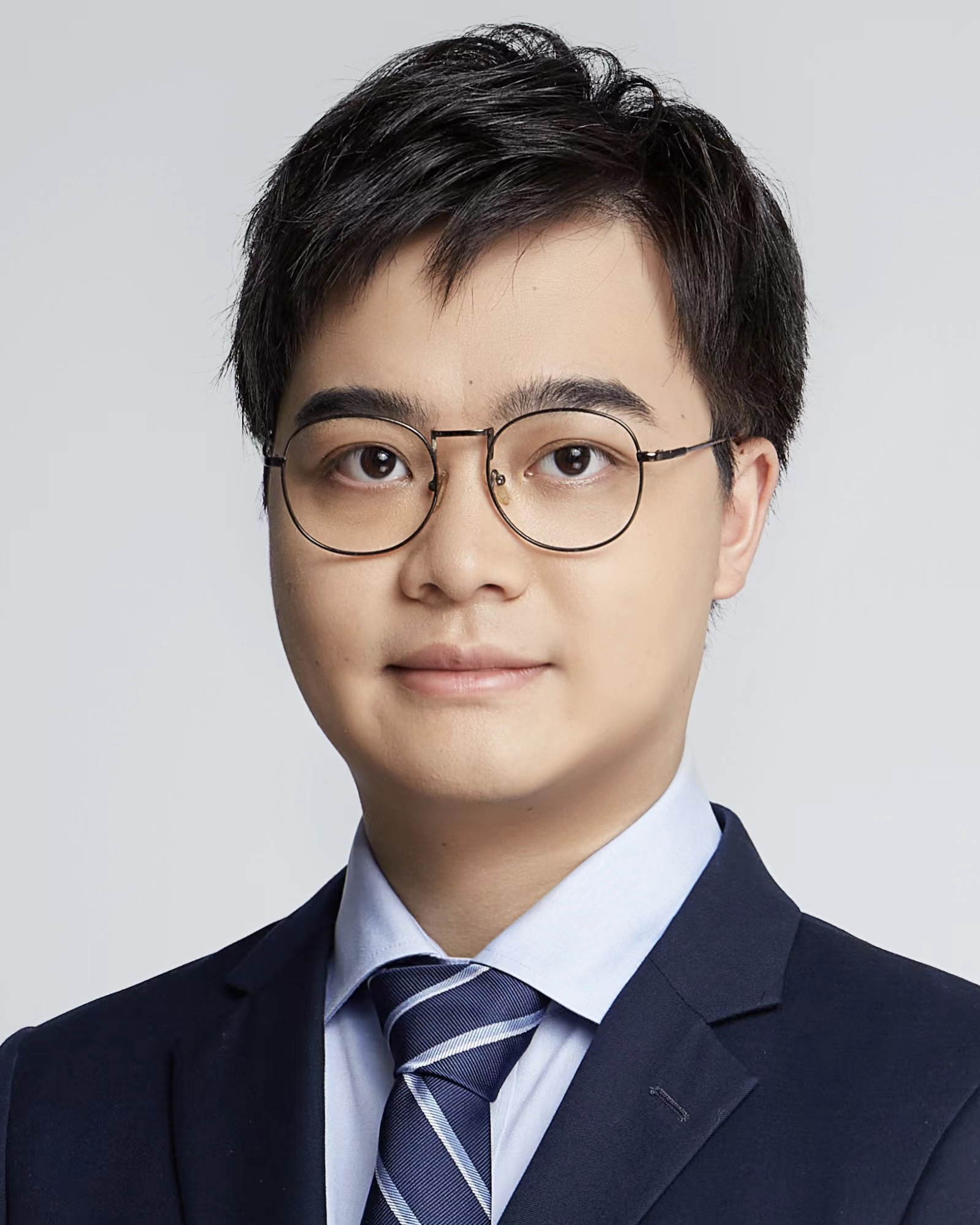}}]{Zhiyuan Xu}
(Member, IEEE) received his Ph.D. degree in computer information science and engineering from Syracuse University, Syracuse, NY, USA, in 2021, and his B.E. degree in the School of Computer Science and Engineering from the University of Electronic Science and Technology of China, Chengdu, China, in 2015. He is now with the AI Innovation Center, Midea Group. His research interests include Deep Learning, Deep Reinforcement Learning, and Robot Learning.
\end{IEEEbiography}

\begin{IEEEbiography}[{\includegraphics[width=1in,height=1.25in, clip, keepaspectratio]{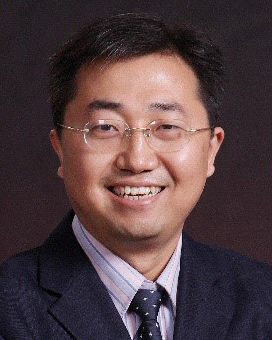}}]{Xiuquan Qiao}
  is currently a Full Professor at the Beijing University of Posts and Telecommunications, Beijing, China, where he is also the Deputy Director of the Key Laboratory of Networking and Switching Technology, Network Service Foundation Research Center of State. He has authored or co-authored over 60 technical papers in international journals and at conferences, including the IEEE Communications Magazine, Proceedings of IEEE, Computer Networks, IEEE Internet Computing, the IEEE Transactions On Automation Science and Engineering, and the ACM SIGCOMM Computer Communication Review. His current research interests include the future Internet, services computing, computer vision, distributed deep learning, augmented reality, virtual reality, and 5G networks. Dr. Qiao was a recipient of the Beijing Nova Program in 2008 and the First Prize of the 13th Beijing Youth Outstanding Science and Technology Paper Award in 2016. He served as the associate editor for the magazine Computing (Springer) and the editor board of China Communications Magazine.
\end{IEEEbiography}

\begin{IEEEbiography}[{\includegraphics[width=1in,height=1.25in, clip, keepaspectratio]{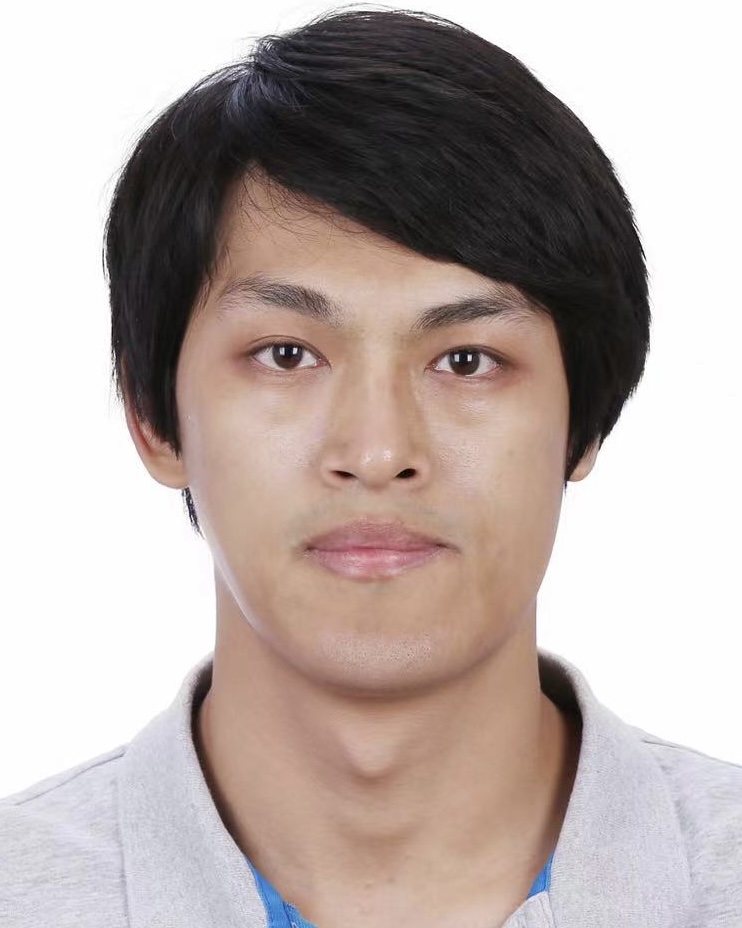}}]{Mengshi Qi}
(Member, IEEE) is currently a professor with the Beijing University of Posts and Telecommunications, Beijing, China. He received the B.S. degree from the Beijing University of Posts and Telecommunications, Beijing, China, in 2012, and the M.S. and Ph.D. degrees in computer science from Beihang University, Beijing, China, in 2014 and 2019, respectively. He  was a postdoctoral researcher with the CVLAB, EPFL, Switzerland from 2019 to 2021.  His research interests include machine learning and computer vision, especially scene understanding, 3D reconstruction, and multimedia analysis.
\end{IEEEbiography}

\begin{IEEEbiography}[{\includegraphics[width=1in,height=1.25in, clip, keepaspectratio]{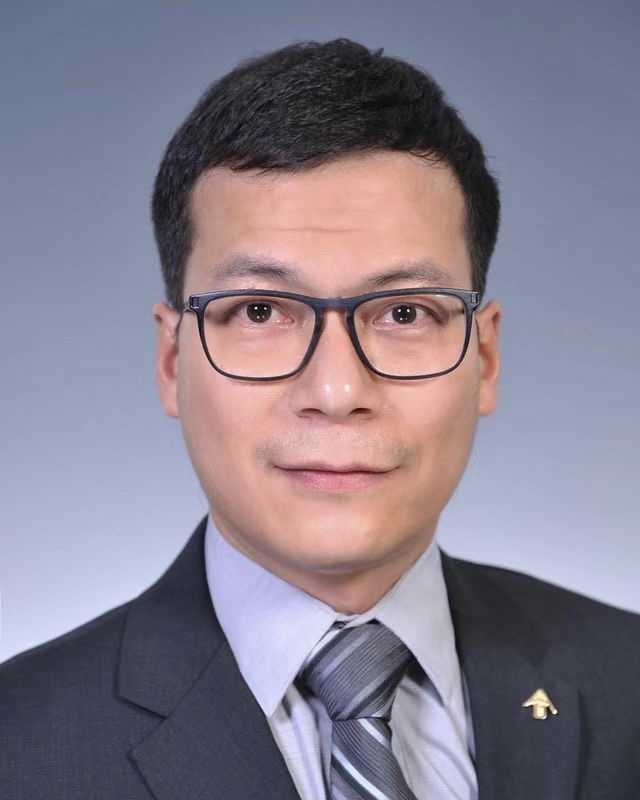}}]{Feifei Feng}
received his B.S. and Ph.D. degrees in electronic engineering from Tsinghua University in 1999 and 2004, respectively.
He is now with the AI Innovation Center, Midea Group.
His research interests include Internet of Things, artificial intelligence and ambient intelligence applications in smart home, and home service robotics.
\end{IEEEbiography}

\begin{IEEEbiography}[{\includegraphics[width=1in,height=1.25in, clip, keepaspectratio]{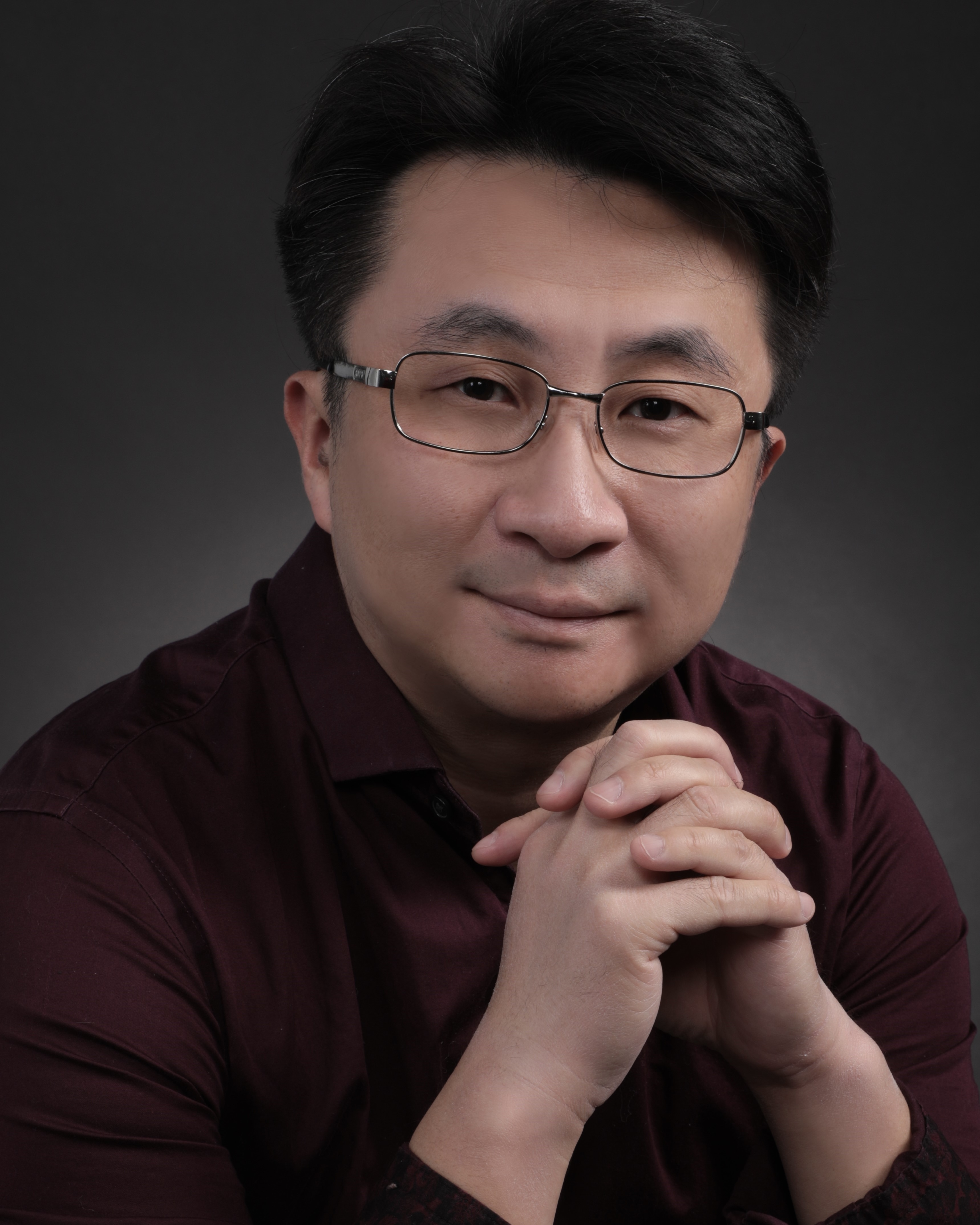}}]{Jian Tang}
(Fellow, IEEE) received his Ph.D. degree in Computer Science from Arizona State University in 2006. He is an IEEE Fellow and an ACM Distinguished Member. He is with Midea Group. His research interests lie in the areas of AI, IoT, Wireless Networking, Mobile Computing and Big Data Systems. He has published over 180 papers in premier journals and conferences. He received an NSF CAREER award in 2009. He also received several best paper awards, including the 2019 William R. Bennett Prize and the 2019 TCBD (Technical Committee on Big Data), Best Journal Paper Award from IEEE Communications Society (ComSoc), the 2016 Best Vehicular Electronics Paper Award from IEEE Vehicular Technology Society (VTS), and Best Paper Awards from the 2014 IEEE International Conference on Communications (ICC) and the 2015 IEEE Global Communications Conference (Globecom) respectively. He has served as an editor for several IEEE journals, including IEEE Transactions on Big Data, IEEE Transactions on Mobile Computing, etc. In addition, he served as a TPC co-chair for a few international conferences, including the IEEE/ACM IWQoS’2019, MobiQuitous’2018, IEEE iThings’2015. etc.; as the TPC vice chair for the INFOCOM’2019; and as an area TPC chair for INFOCOM 2017- 2018. He is also an IEEE VTS Distinguished Lecturer, and the Chair of the Communications Switching and Routing Committee of IEEE ComSoc 2020-2021.
\end{IEEEbiography}

\end{document}